\documentclass{article}

\PassOptionsToPackage{numbers,sort&compress}{natbib}

\usepackage[main,final]{neurips_2026}

\usepackage[utf8]{inputenc} %
\usepackage[T1]{fontenc}    %
\usepackage{hyperref}       %
\usepackage{url}            %
\usepackage{booktabs}       %
\usepackage{amsfonts}       %
\usepackage{nicefrac}       %
\usepackage{microtype}      %
\usepackage{xcolor}         %
\usepackage{graphicx}
\usepackage{subcaption}
\usepackage{amsmath}
\usepackage{amssymb}
\usepackage{mathtools}
\usepackage{amsthm}
\usepackage{algorithm}
\usepackage{algcompatible}
\usepackage{placeins}
\usepackage[capitalize,noabbrev]{cleveref}

\makeatletter
\@ifundefined{theHALG@line}
  {\newcommand{\theHALG@line}{\thealgorithm.\arabic{ALG@line}}}
  {\renewcommand{\theHALG@line}{\thealgorithm.\arabic{ALG@line}}}
\makeatother

\theoremstyle{plain}

\theoremstyle{definition}

\theoremstyle{remark}

\title{BridgeDock: Taming Generative Co-Folding Prior for Molecular Docking with Diffusion Bridge}

\author{%
  \textbf{Shikun Feng}$^{1,2}$,\quad
  \textbf{Zipeng Yang}$^{1,3}$,\quad
  \textbf{Qiuyi Li}$^{1,2}$,\quad
  \textbf{Mingze Yin}$^{4}$ \\
  \textbf{Xiaoyuan Zhang}$^{1,2}$,\quad
  \textbf{Tingjun Hou}$^{5}$,\quad
  \textbf{Yiheng Zhu}$^{1,2}$\thanks{Corresponding author: \href{mailto:zhuyiheng@zgci.ac.cn}{\texttt{zhuyiheng@zgci.ac.cn}}.} \\[0.5em]
  $^{1}$Zhongguancun Academy, Beijing, China. \\
  $^{2}$Zhongguancun Institute of Artificial Intelligence, Beijing, China. \\
  $^{3}$School of Physical Science and Technology, \\
  Beijing University of Posts and Telecommunications, Beijing, China. \\
  $^{4}$College of Computer Science and Technology, Zhejiang University, Hangzhou, China. \\
  $^{5}$College of Pharmaceutical Sciences, Zhejiang University, Hangzhou, China.
}

\begin{document}

\maketitle

\begin{abstract}

Molecular docking aims to predict the three-dimensional structure of a protein–ligand complex; in the conventional structure-based setting, it is conditioned on an input protein structure and ligand, often starting from an apo receptor in flexible docking or a holo receptor in rigid docking. Recent co-folding models exemplified by AlphaFold 3 achieve strong accuracy in protein–ligand complex prediction, but they typically generate complexes from sequence-derived inputs and therefore do not directly use apo structures as structural priors in docking, leading to a larger conformational search space and weaker robustness in difficult or few-step settings. We present BridgeDock, a framework that adapts pretrained co-folding backbones to flexible docking by modeling the apo-to-holo transition as a diffusion bridge, allowing generation from the initial protein–ligand state rather than from Gaussian noise. To make this bridge formulation compatible with pretrained denoisers, we introduce an alignment mechanism that maps bridge states to the original denoising schedule of the pretrained model. Experiments on standard flexible docking benchmarks show that BridgeDock consistently outperforms strong co-folding baselines. Moreover, BridgeDock is computationally efficient at inference time and remains effective even with very few denoising steps.

\end{abstract}

\section{Introduction}

Molecular docking aims to predict the three-dimensional structure of a complex formed by a small molecule and its binding protein. Because the resulting structure can be used to infer protein-ligand binding modes and underlying interaction mechanisms, docking is a cornerstone of structure-based drug design. With advances in deep learning, a range of learning-based docking methods has emerged, including regression-based approaches~\cite{lu2022tankbind, fengfigrdock} and generative approaches~\cite{stark2022equibind,plainer2023diffdock}. These methods are generally more efficient than traditional docking pipelines~\cite{trott2010autodock,halgren2004glide}, which rely on scoring-function--guided search and iterative optimization to identify the best-ranked poses.

Recent co-folding approaches, exemplified by AlphaFold 3~\cite{abramson2024alphafold3} and its open-source reproductions or AF3-style implementations, such as Boltz-1~\cite{wohlwend2024boltz1} and Protenix~\cite{chen2025protenix}, benefit from training on large-scale biological data and from sophisticated network designs, achieving very high predictive accuracy for protein monomers as well as diverse complexes, including protein--protein, protein--nucleic acid, and protein-ligand assemblies. On molecular docking tasks, these methods can substantially outperform earlier deep learning--based docking approaches. However, their problem formulation differs from the standard docking setup. In conventional molecular docking, the initial structures of the protein and ligand are typically provided: rigid docking usually assumes a holo receptor and focuses on predicting the ligand pose, whereas flexible docking often starts from an apo receptor and requires modeling conformational changes in both the ligand and the protein. In contrast, co-folding methods generate complexes from sequence-derived inputs and therefore do not directly use apo structures as structural priors in docking.

Due to limited training data and the reliance on relatively homogeneous protein-ligand datasets, deep learning--based docking methods still exhibit a substantial performance gap compared with co-folding approaches, even when initial structures are provided. This gap raises a key question: \textbf{can we leverage the strong priors learned by co-folding models for molecular docking, while explicitly incorporating initial structural information, to further push the upper bound of docking accuracy?}

\begin{figure}[t]
    \centering
    \includegraphics[width=0.8\linewidth]{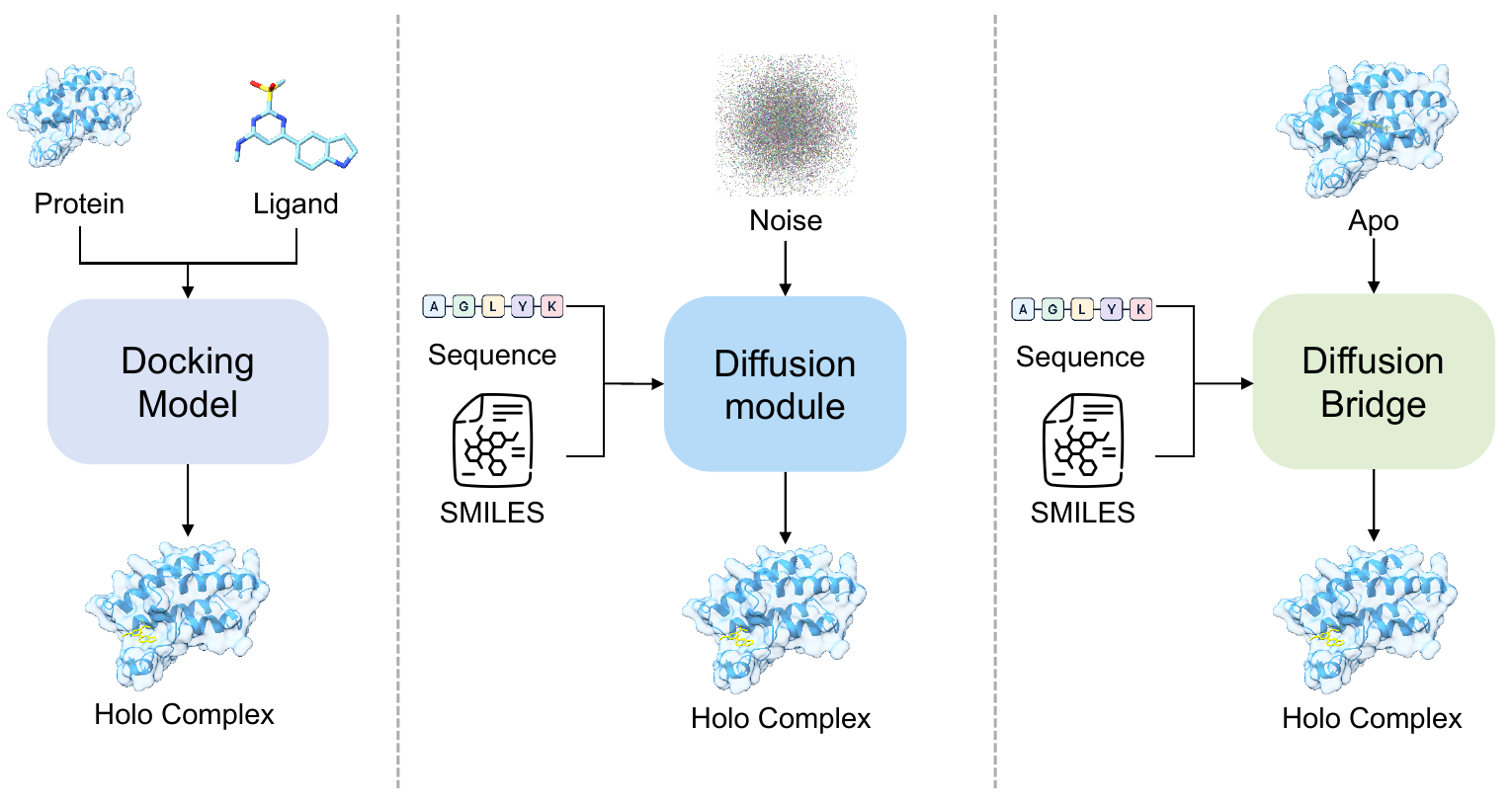}
    \caption{
    Comparison of three paradigms for protein-ligand docking. \textbf{Left:} Traditional deep learning-based methods directly predict holo complexes from protein and ligand inputs. \textbf{Middle:} Recent co-folding methods formulate docking as a generation process starting from noise. \textbf{Right:} BridgeDock leverages apo structures as explicit structural priors and formulates docking as an apo-to-holo diffusion bridge.
    }
    \label{fig:intro_overview}
\end{figure}

AlphaFold~3 formulates structure prediction as a denoising diffusion process: during training, the model learns to map noise-corrupted atomic coordinates back to their corresponding clean (noise-free) coordinate representations; during inference, it conditions on sequence-derived features produced by frontend pipelines (e.g., MSAs and, optionally, structural templates) and progressively generates 3D atomic coordinates from Gaussian noise, following a noise-to-data paradigm. In this work, we focus on the more challenging flexible docking setting, where the starting point consists of unbound protein and ligand structures rather than pure noise. Instead of treating the initial structures as simple conditioning inputs, we adopt the diffusion bridge~\cite{zhou2024denoising} to construct a stochastic process between the distribution of initial structures and that of the target bound complexes. Compared with standard diffusion, this formulation more naturally models the continuous transition from unbound to bound states and is expected to improve both generation accuracy and efficiency. A conceptual comparison between traditional docking methods, co-folding approaches, and our proposed BridgeDock framework is illustrated in Figure~\ref{fig:intro_overview}.

However, directly training a diffusion bridge leads to a mismatch with the original diffusion formulation used in prior co-folding methods, which prevents effective reuse of model parameters pre-trained on large-scale and diverse datasets, naive fine-tuning in this setting can even result in capability degradation. Motivated by recent alignment strategies explored in video generation~\cite{wang2024framebridge}, we introduce a mechanism that reparameterizes intermediate diffusion-bridge states—composed of the initial structure, the target complex, and injected noise—into the standard diffusion formulation. Furthermore, we map time steps across the stochastic paths of the diffusion bridge and the standard diffusion process, enforcing an alignment such that corresponding states exhibit the same signal-to-noise ratio. This design makes bridge-based docking compatible with pretrained co-folding denoisers and allows us to better reuse their learned structural priors in the docking setting.

Experimental results demonstrate that BridgeDock not only outperforms sequence-based co-folding model baselines, but also exceeds baselines that apply supervised fine-tuning using apo structures as direct inputs. These results indicate that BridgeDock effectively leverages co-folding priors while further enhancing model performance on molecular docking tasks. Moreover, generating structures from informed initial states provides a stronger conditioning signal, enabling faster convergence to the target complex and higher efficiency than denoising from pure noise. As a result, BridgeDock exhibits superior sampling efficiency and achieves competitive performance even in the one-step generation setting. Beyond molecular docking, our framework offers broader insights for general structure prediction and other data-to-data generative problems.

\section{Related Work}

\subsection{Molecular Docking}

\textbf{Structure-based Molecular Docking} Traditional molecular docking methods~\cite{friesner2004glide,morris1996distributed,trott2010autodock} typically formulate docking as conformational search coupled with empirical scoring functions. Early neural docking methods replace this search-and-score pipeline with regression-based geometric prediction: EquiBind~\cite{stark2022equibind} and E3Bind~\cite{zhang2023e3bind} directly regress docked complex coordinates, while TankBind~\cite{lu2022tankbind} predicts docking poses and binding affinities in a multitask framework. Despite their efficiency, such deterministic models provide a single-shot prediction and are less suited to capturing the distributional nature of docking, where multiple plausible binding poses may exist for the same protein-ligand pair. DiffDock~\cite{corso2023diffdock} addresses this limitation by formulating ligand docking as a generative diffusion process over ligand translation, rotation, and torsions. This generative paradigm has since been extended to flexible docking: DiffDock-Pocket~\cite{plainer2023diffdock}, DiffBindFR~\cite{zhu2024diffbindfr}, and Re-Dock~\cite{huang2024re} jointly model ligand poses and pocket side-chain conformational changes, whereas FlexDock~\cite{corsocomposing} and DynamicBind~\cite{lu2024dynamicbind} further account for broader protein flexibility, including backbone-level conformational changes.

\textbf{Co-folding Models for Complex Prediction} Beyond docking-specific methods, recent co-folding and general-purpose biomolecular structure prediction models have sought to predict protein-ligand complex structures directly from protein sequence and ligand molecular representations. NeuralPLexer~\cite{qiao2024state} predicts protein-ligand complex structures from protein sequences and ligand molecular graphs with a generative model, while Umol~\cite{bryant2024structure} performs all-atom protein-ligand co-folding directly from sequence-level information. More recently, AlphaFold 3~\cite{abramson2024alphafold3} substantially generalizes this paradigm by training a diffusion-based architecture on large-scale structural data to predict the joint structures of biomolecular complexes, including proteins, nucleic acids, small molecules, ions, and modified residues. It achieves strong accuracy across diverse biomolecular interaction types, including protein-ligand complexes.

\subsection{Diffusion Bridges}

The Schr\"odinger bridge (SB) problem, a classical entropy‑regularized optimal transport formulation~\citep{schrodinger1932theorie, leonard2014survey, chen2021optimal}, seeks a stochastic process closest to a reference (e.g., Brownian motion) while matching prescribed marginals. This framework naturally connects to diffusion-based generative models~\citep{vargas2021solving, wang2021deep, de2021diffusion, shi2024diffusion}, where the prior is often Gaussian noise~\citep{song2021scorebased}, and has been extended to approximate arbitrary reference processes from coupled samples~\citep{holdijk2022path, somnath2023aligned, liu2023i2sb, zhou2024denoising}. Recent work on Markov bridges~\citep{fitzsimmons1992markovian, igashov2024retrobridge, zhu2024bridge,schneuing2025multidomain} generalizes SB to categorical and multimodal distributions. In this work, we make the first attempt to synergize the theoretical advantages of diffusion bridge models with the powerful generative priors of co‑folding methods to improve molecular docking.

\section{Methods}

\subsection{AF3 Diffusion and Docking Setup}
\label{sec:af3_diffusion}

Molecular docking aims to predict the bound structure of a
protein-ligand complex. As shown in the upper part of
Figure~\ref{fig:main_overview}, AlphaFold~3 (AF3) approaches this task through a
noise-to-structure diffusion paradigm: it initializes atomic coordinates from
Gaussian noise and iteratively denoises them into the final complex structure
under sequence-derived conditioning representations.We follow its EDM-style
preconditioned denoising view and
write $\mathbf{x}=\{\mathbf{x}_l\}_{l=1}^L$ for all heavy-atom coordinates,
$\mathbf{x}^{\mathrm{apo}}$ for the apo-conditioned initialization, and
$\mathbf{x}^{\mathrm{GT}}$ for the target holo structure. We denote the AF3
input features and trunk representations by
$\mathcal{C}=(\{f^*\},\{s_i^{\mathrm{inputs}}\},
\{s_i^{\mathrm{trunk}}\},\{z_{ij}^{\mathrm{trunk}}\})$.

During AF3 training, the noise level $\hat{t}$ is sampled from a log-normal
distribution and independent Gaussian noise is added to the target coordinates:
\begin{equation}
\boldsymbol{\epsilon} \sim \mathcal{N}(\mathbf{0}, I),
\qquad
\mathbf{x}^{\mathrm{noisy}}
=
\mathbf{x}^{\mathrm{GT}} + \hat{t}\,\boldsymbol{\epsilon} ,
\end{equation}
where $\boldsymbol{\epsilon}$ is the coordinate noise. The exact sampling rule
for $\hat{t}$ is given in \Cref{app:af3_sampling}.

Using the notation introduced above, the preconditioned diffusion module can be
written in the EDM-style form
\begin{equation}
\hat{\mathbf{x}}_0
=
c_{\mathrm{skip}}(\hat{t})\,\mathbf{x}^{\mathrm{noisy}}
+
c_{\mathrm{out}}(\hat{t})\,
F_{\theta}\!\left(
c_{\mathrm{in}}(\hat{t})\,\mathbf{x}^{\mathrm{noisy}},
\mathcal{C}
\right).
\end{equation}
Here, $\sigma_{\mathrm{data}}$ is the data-scale hyperparameter, and the
preconditioning coefficients are
\begin{equation}
c_{\mathrm{in}}(\hat{t})
=
\frac{1}{\sqrt{\hat{t}^2+\sigma_{\mathrm{data}}^2}},
\qquad
c_{\mathrm{skip}}(\hat{t})
=
\frac{\sigma_{\mathrm{data}}^2}{\sigma_{\mathrm{data}}^2+\hat{t}^2},
\qquad
c_{\mathrm{out}}(\hat{t})
=
\frac{\sigma_{\mathrm{data}}\hat{t}}
{\sqrt{\hat{t}^2+\sigma_{\mathrm{data}}^2}}.
\end{equation}
The function $F_{\theta}$ is the residual coordinate network: it takes the
scaled noisy coordinates and conditioning context as input, and its output is
multiplied by $c_{\mathrm{out}}(\hat{t})$. For compactness, we denote the
complete preconditioned denoiser by
$D_{\theta}(\mathbf{x}^{\mathrm{noisy}},\hat{t},\mathcal{C})$, while
$F_{\theta}$ denotes the residual network inside this preconditioning wrapper.

In the training loss, AF3 first rigidly aligns the ground-truth structure to
the denoised prediction and then minimizes the weighted aligned reconstruction
loss
\begin{equation}
\mathcal{L}_{\mathrm{diffusion}}
=
\frac{\hat{t}^2+\sigma_{\mathrm{data}}^2}
{(\hat{t}+\sigma_{\mathrm{data}})^2}
\left\|
\hat{\mathbf{x}}_{0} - \mathbf{x}^{\mathrm{GT\mbox{-}aligned}}
\right\|^2.
\end{equation}
Here, $\{\mathbf{x}^{\mathrm{GT\mbox{-}aligned}}_l\}$ denotes the ground-truth
coordinates after weighted rigid alignment to the denoised prediction.

In the original AF3 formulation, this loss further includes several
task-specific refinements, such as atom-type-dependent reweighting in the
aligned reconstruction term and additional bond-geometry constraints for
bonded ligands or glycans. We omit these terms in the definition of the loss
functions for simplicity, since they are not essential to the presentation of
the core generative process.

At inference time, AF3 follows a discrete reverse diffusion schedule from
Gaussian initialization to final all-atom coordinates. Pseudocode and
hyperparameters are provided in \Cref{alg:af3_sampling,app:af3_sampling}.
Although this noise-to-structure formulation benefits from the structural
knowledge learned by AF3 during large-scale co-folding pretraining, it does not directly exploit the provided apo structure as a structural
prior in the docking setting. This motivates the apo-to-holo bridge formulation
introduced next.

\subsection{BridgeDock: Apo-to-Holo Diffusion Bridge}
\label{sec:bridge_diffusion}

\begin{figure}[t]
    \centering
    \includegraphics[width=\linewidth,
     trim=0 0 0 0.5cm,
     clip
    ]{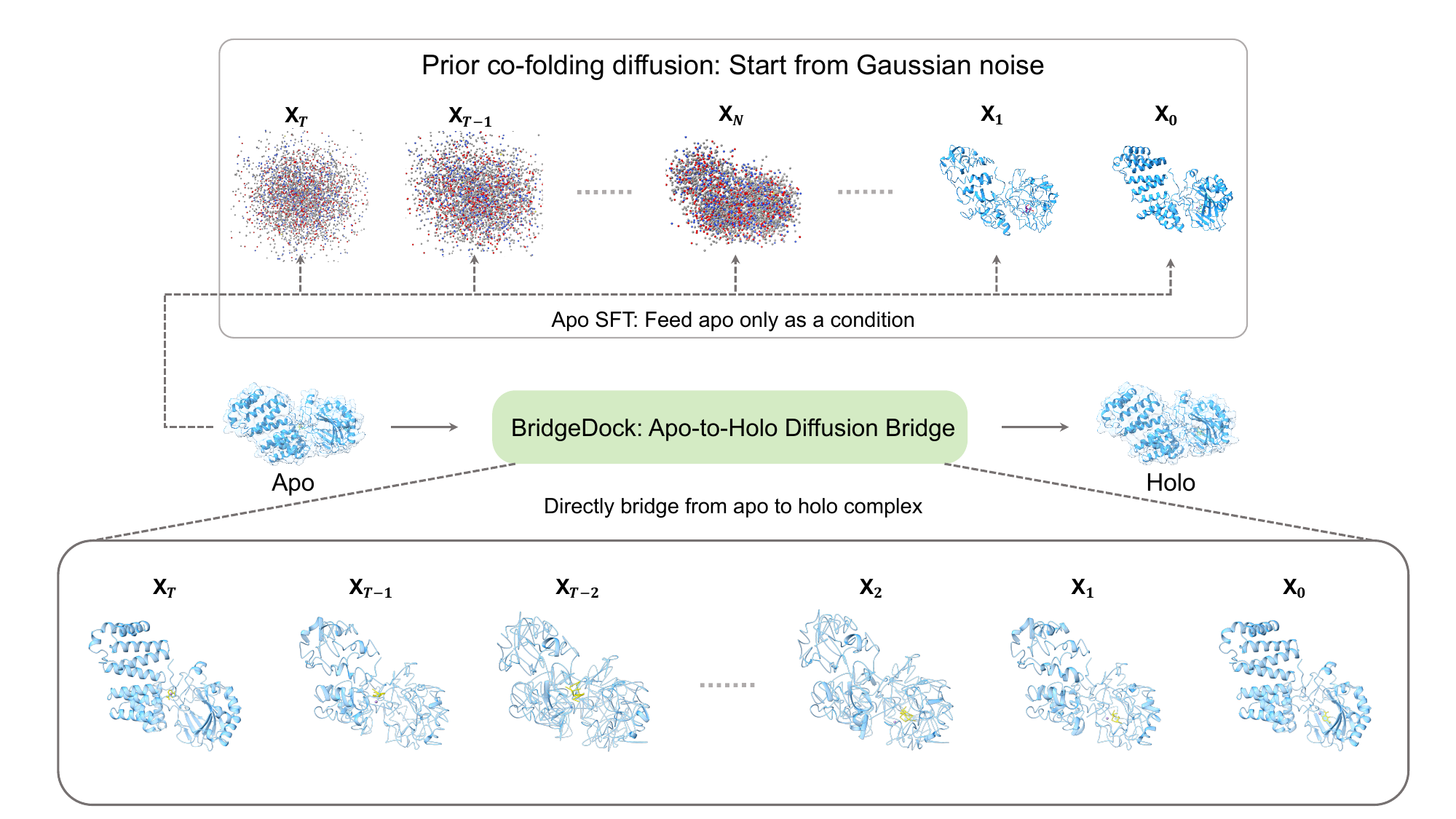}
    \caption{
    Overview of BridgeDock. Unlike prior co-folding diffusion methods that start from Gaussian noise, and Apo SFT which only conditions on apo structures, BridgeDock formulates protein-ligand docking as an apo-to-holo diffusion bridge, directly transforming apo conformations into target holo complexes.
    }
    \label{fig:main_overview}
\end{figure}

As illustrated in Figure~\ref{fig:main_overview}, BridgeDock replaces AF3's noise-to-data generation process with an apo-to-holo
diffusion bridge. The reverse process starts from the apo-conditioned initial
structure $\mathbf{x}^{\mathrm{apo}}$, which contains the apo
protein coordinates and an initial ligand placement, and moves toward the holo
target structure. By introducing this diffusion bridge, we explicitly model the
continuous transition from the apo state to the holo state during generation,
which helps improve both generation accuracy and sampling efficiency.

Following the variance-exploding (VE) formulation of Denoising Diffusion Bridge
Models (DDBM)~\cite{zhou2024denoising}, we define the forward bridge state as
\begin{equation}
\mathbf{x}_{t}
=
a_t \mathbf{x}^{\mathrm{apo}}
+
b_t \mathbf{x}^{\mathrm{GT}}
+
c_t \boldsymbol{\epsilon},
\qquad
\boldsymbol{\epsilon} \sim \mathcal{N}(\mathbf{0}, I),
\label{eq:bridge_forward}
\end{equation}
with coefficients
\begin{equation}
a_t = \frac{t^2}{\sigma_{\max}^2},
\qquad
b_t = 1 - \frac{t^2}{\sigma_{\max}^2},
\qquad
c_t = t \sqrt{1 - \frac{t^2}{\sigma_{\max}^2}}.
\label{eq:bridge_abc}
\end{equation}
Here $t\in(\sigma_{\min},\sigma_{\max})$ is the bridge time sampled during
training, and $\boldsymbol{\epsilon}$ is Gaussian coordinate noise. At
$t=\sigma_{\max}$, the bridge is anchored at the apo structure because
$a_t=1,b_t=c_t=0$; as $t$ approaches zero, the deterministic component
approaches the holo target. Thus the noisy state $\mathbf{x}_t$ explicitly
couples the apo starting structure with the holo target, rather than perturbing
only $\mathbf{x}^{\mathrm{GT}}$ as in AF3.

For the raw bridge parameterization, the corresponding DDBM-style preconditioned
denoiser predicts the holo endpoint through a skip connection from the current
bridge state plus a scaled neural-network residual:
\begin{equation}
\hat{\mathbf{x}}_{0}
=
c_{\mathrm{skip}}(t)\,\mathbf{x}_{t}
+
c_{\mathrm{out}}(t)\,
F_{\theta}\!\left(
c_{\mathrm{in}}(t)\,\mathbf{x}_{t},
\mathbf{x}^{\mathrm{apo}},
\mathcal{C}
\right),
\label{eq:bridge_denoiser}
\end{equation}
where $\hat{\mathbf{x}}_{0}$ denotes the predicted clean endpoint
corresponding to the target holo structure.
In addition to defining the bridge endpoint, $\mathbf{x}^{\mathrm{apo}}$ is
retained as an explicit structural condition of the denoiser throughout training
and inference. For compactness, we denote the complete preconditioned denoising
pipeline by
$D_{\theta}(\mathbf{x}_{t},t,\mathbf{x}^{\mathrm{apo}},\mathcal{C})$.
This notation includes the apo encoder that maps the raw apo coordinates into
network-conditioning features.

To define the preconditioning coefficients, let
\begin{equation}
A_t
=
a_t^2 \sigma_{\mathrm{data,end}}^2
+
b_t^2 \sigma_{\mathrm{data}}^2
+
2 a_t b_t \,\mathrm{cov}_{xy}
+
\kappa^2 c_t^2.
\end{equation}
Then
\begin{equation}
c_{\mathrm{in}}(t)=\frac{1}{\sqrt{A_t}},
\end{equation}
\begin{equation}
c_{\mathrm{skip}}(t)
=
\frac{
b_t \sigma_{\mathrm{data}}^2 + a_t \,\mathrm{cov}_{xy}
}{A_t},
\end{equation}
and
\begin{equation}
c_{\mathrm{out}}(t)
=
\frac{
\sqrt{
a_t^2\!\left(
\sigma_{\mathrm{data,end}}^2 \sigma_{\mathrm{data}}^2
-
\mathrm{cov}_{xy}^2
\right)
+
\sigma_{\mathrm{data}}^2 \kappa^2 c_t^2
}
}{\sqrt{A_t}}.
\end{equation}
These coefficients match the generalized DDBM preconditioning specialized to the
VE bridge. The hyperparameters $\sigma_{\mathrm{data}}$,
$\sigma_{\mathrm{data,end}}$, $\mathrm{cov}_{xy}$, and $\kappa$ are listed in
\Cref{app:hyperparameters}.

The same preconditioning induces a bridge-time reconstruction weight,
\begin{equation}
w_{\mathrm{bridge}}(t)
=
\frac{1}{c_{\mathrm{out}}(t)^2}.
\end{equation}
We apply this weight to the aligned denoising objective introduced in
\Cref{sec:af3_aligned_bridge}. Other auxiliary structure losses, such as
smooth-LDDT and optional bond-geometry terms, follow the AF3 objective.

For inference, we adapt the Diffusion Bridge Implicit Models (DBIM)
sampler~\cite{zheng2024diffusion} to our apo-to-holo VE bridge. Instead of
starting from Gaussian coordinates as in AF3, the reverse process is initialized
from the apo-conditioned initial structure $\mathbf{x}^{\mathrm{apo}}$ and then follows a
DBIM-style implicit bridge trajectory toward the holo target. The reverse
update uses the bridge coefficients above; the concrete time grid,
initialization, reverse update, and pseudocode are deferred to
\Cref{app:bridge_inference_details}.

\subsection{AF3-Aligned Bridge Parameterization}
\label{sec:af3_aligned_bridge}

Our goal is to adapt the pretrained AF3 denoiser into an apo-to-holo bridge
generator, so that it can start from the provided docking structure while still
exploiting the structural prior learned from large-scale co-folding. A naive
bridge objective creates a tension: the model is expected to reuse its
pretrained generative capability, yet the raw bridge states change both the
coordinate parameterization and the noise-level interface on which that
capability was learned. This mismatch makes optimization substantially harder
and can weaken the pretrained prior. BridgeDock therefore keeps the bridge
forward process unchanged, but reparameterizes each interior bridge state as an
AF3-style noisy target before applying the denoiser.

The VE bridge state in \Cref{eq:bridge_forward} depends on both
$\mathbf{x}^{\mathrm{apo}}$ and $\mathbf{x}^{\mathrm{GT}}$. In contrast, AF3 is
trained on noisy targets of the form
\begin{equation}
\mathbf{x}^{\mathrm{noisy}}
=
\mathbf{x}^{\mathrm{GT}} + \sigma \boldsymbol{\epsilon},
\end{equation}
where $\sigma$ is the denoiser's noise-level input. The alignment below converts
the bridge state at time $t$ into this AF3-compatible form.

For $b_t>0$, define the aligned noisy coordinate as
\begin{equation}
\mathbf{x}^{\mathrm{align}}_{t}
=
\frac{\mathbf{x}_{t} - a_t \mathbf{x}^{\mathrm{apo}}}{b_t}.
\end{equation}
Substituting the bridge expression yields
\begin{align}
\mathbf{x}^{\mathrm{align}}_{t}
&=
\frac{
a_t \mathbf{x}^{\mathrm{apo}}
+
b_t \mathbf{x}^{\mathrm{GT}}
+
c_t \boldsymbol{\epsilon}
-
a_t \mathbf{x}^{\mathrm{apo}}
}{b_t} \\
&=
\mathbf{x}^{\mathrm{GT}}
+
\frac{c_t}{b_t}\boldsymbol{\epsilon}.
\end{align}
Therefore, the aligned variable has the AF3 denoising form
\begin{equation}
\mathbf{x}^{\mathrm{align}}_{t}
=
\mathbf{x}^{\mathrm{GT}} + \sigma^{\mathrm{align}}_t \boldsymbol{\epsilon},
\qquad
\sigma^{\mathrm{align}}_t = \frac{c_t}{b_t}.
\label{eq:aligned_noisy_target}
\end{equation}
The denoiser uses
$(\mathbf{x}^{\mathrm{align}}_t,\sigma^{\mathrm{align}}_t)$ as its coordinate
and noise-level inputs, replacing the bridge time $t$ with the AF3-compatible
noise level. The apo encoder separately maps
$\mathbf{x}^{\mathrm{apo}}$ into conditioning features, so removing the apo
term from the coordinate input does not remove apo conditioning.

This mapping preserves the relative target-to-noise scale of the bridge state:
\begin{equation}
\mathrm{SNR}_{\mathrm{bridge}}(t)
\propto
\frac{b_t^2}{c_t^2},
\qquad
\mathrm{SNR}_{\mathrm{align}}(t)
\propto
\frac{1}{(\sigma^{\mathrm{align}}_t)^2}
=
\frac{b_t^2}{c_t^2}.
\end{equation}
Thus, the alignment preserves the bridge SNR relative to the target endpoint
while specifying the AF3 noise level seen by the denoiser. The clean endpoint is
predicted as
\begin{equation}
\hat{\mathbf{x}}_{0}
=
D_{\theta}\!\left(
\mathbf{x}^{\mathrm{align}}_{t},
\sigma^{\mathrm{align}}_t,
\mathbf{x}^{\mathrm{apo}},
\mathcal{C}
\right).
\end{equation}
The final bridge loss applies the bridge-time weight from
\Cref{sec:bridge_diffusion} to this AF3-aligned prediction:
\begin{equation}
\mathcal{L}_{\mathrm{bridge}}
=
w_{\mathrm{bridge}}(t)
\left\|
\hat{\mathbf{x}}_{0}
-
\mathbf{x}^{\mathrm{GT\mbox{-}aligned}}
\right\|^2.
\end{equation}
Consequently, the denoising target seen by the network remains close to the
original AF3 training target, while the generative process remains an
apo-to-holo bridge.

The same reparameterization is used during DBIM inference. For an interior
bridge state $\mathbf{x}_s$ with $b_s>0$, where $a_s,b_s,c_s$ are the VE bridge
coefficients in \Cref{eq:bridge_abc}, we compute
\begin{equation}
\mathbf{x}^{\mathrm{align}}_s
=
\frac{\mathbf{x}_s - a_s \mathbf{x}^{\mathrm{apo}}}{b_s},
\qquad
\sigma^{\mathrm{align}}_s = \frac{c_s}{b_s},
\end{equation}
and apply the denoiser as
\begin{equation}
\hat{\mathbf{x}}_0
=
D_{\theta}\!\left(
\mathbf{x}^{\mathrm{align}}_s,
\sigma^{\mathrm{align}}_s,
\mathbf{x}^{\mathrm{apo}},
\mathcal{C}
\right).
\end{equation}
This estimate is then used in the DBIM reverse bridge update. In our
AF3-aligned implementation, sampling starts from a near-apo interior time rather
than the singular endpoint $b_{\sigma_{\max}}=0$, and the aligned coordinate is
centered before denoising. Alignment changes only the denoiser interface;
the reverse process retains the same explicit apo condition at every step
and moves toward the holo target manifold.

Detailed training and AF3-aligned DBIM inference procedures are provided in
\Cref{alg:af3_aligned_bridge_train,alg:af3_aligned_bridge_test}
(\Cref{app:af3_aligned_algorithms}).

\section{Experiments}

\subsection{Main Experiments}

\paragraph{Experimental setup}

For co-folding models, we use Protenix~\cite{chen2025protenix}, an
open-source AF3-style implementation, as the baseline family. Specifically, we
use its lightweight Protenix-Mini variant~\cite{gong2025protenixmini} as the
backbone for continued training. We train BridgeDock with the AF3-aligned bridge
objective on 40,000 protein-ligand complexes curated from the original Protenix
training data for 80,000 steps. Training is performed on 8$\times$A100 GPUs and
takes approximately five days.

We evaluate on PoseBusters v2~\cite{buttenschoen2024posebusters} (hereafter
PoseBusters) and a time-based test split constructed from PDBBind
v2020~\cite{liu2015pdbbind}. For apo-conditioned inputs, protein conformations
are predicted from amino-acid
sequences, ligand conformations are generated using RDKit, and the initial
complex is formed by placing the ligand at the geometric center of the apo
protein. Importantly, BridgeDock is not given binding-pocket annotations or
pocket-center information at test time; the ligand is initialized from this
global apo-protein center rather than from an oracle pocket location.
This setup matches the flexible docking scenario motivating BridgeDock: the
model is given an informative but unbound starting structure and must generate
the corresponding bound complex.

\paragraph{Evaluation metrics}
We report five complementary metrics: lDDT, the percentages of predictions whose
ligand RMSD is below 2\,\AA{} and 5\,\AA{}, and the mean and median ligand RMSD.
lDDT measures local structural fidelity in a superposition-free manner, while
the two threshold-based percentages capture ligand pose accuracy at commonly
used RMSD cutoffs. Mean ligand RMSD reflects the overall error magnitude, whereas
median ligand RMSD is less sensitive to rare large failures. Together, these
metrics distinguish improvements in global robustness from improvements on the
typical easy cases.

\paragraph{Baselines}
The primary comparison is against the sequence-conditioned AF3-style baseline
implemented with Protenix-Mini, which uses the same co-folding backbone but does
not initialize the generative trajectory from the apo complex. Unless otherwise
specified, we use 20 denoising steps for both the AF3 baseline and BridgeDock.
For few-step PoseBusters comparisons, we use the ODE sampler proposed in
Protenix-Mini~\cite{gong2025protenixmini} for the AF3 baseline rather than the
original stochastic AF3-style sampler, giving the baseline a strong sampler
under reduced sampling budgets. On PDBBind, we additionally compare with
search-based docking methods, including SMINA and GNINA, and with recent
flexible docking models, including DiffDock-Pocket, Re-Dock, and FlexDock. The
reported results for these docking methods are taken from published benchmark
evaluations under their respective known-pocket protocols. In particular,
DiffDock-Pocket, Re-Dock, and FlexDock use ligand-derived pocket information;
their pocket definitions and protein-cropping rules are detailed in
\Cref{app:docking_baselines}. These comparisons test whether BridgeDock can retain the strong
co-folding prior while making better use of the apo structural information that
is available in conventional docking.

\subsubsection{PoseBusters}

\begin{table}[t]
\centering
\small
\caption{Docking performance on PoseBusters under different denoising budgets.
The 20-step setting is the default used in the main comparison. Higher values
indicate better performance for lDDT and for the percentages of predictions with
ligand RMSD below 2\,\AA{} or 5\,\AA{}, whereas lower values indicate better
performance for mean and median ligand RMSD.}
\setlength{\tabcolsep}{4pt}
\begin{tabular}{@{}lcccccc@{}}
\toprule
& \multicolumn{3}{c}{Docking Accuracy $\uparrow$} & \multicolumn{2}{c}{Ligand RMSD (\AA) $\downarrow$} & \\
\cmidrule(lr){2-4}\cmidrule(lr){5-6}
Model & lDDT (\%) & \%<2 $\uparrow$ & \%<5 $\uparrow$ & Mean & Median & \# Steps \\
\midrule
AF3 baseline (ODE) & 83.91 & 69.76 & 86.51 & 2.33 & 1.19 & 1 \\
BridgeDock & \textbf{90.77} & \textbf{69.86} & \textbf{88.04} & \textbf{2.19} & \textbf{1.08} & 1 \\
\midrule
AF3 baseline (ODE) & 84.26 & 68.23 & 84.40 & 2.72 & 1.14 & 5 \\
BridgeDock & \textbf{91.23} & \textbf{71.77} & \textbf{88.52} & \textbf{2.33} & \textbf{1.04} & 5 \\
\midrule
AF3 baseline & 84.79 & 68.61 & 84.69 & 2.64 & 1.09 & 20 \\
BridgeDock & \textbf{90.41} & \textbf{73.21} & \textbf{89.47} & \textbf{2.26} & 1.09 & 20 \\
\bottomrule
\end{tabular}
\label{tab:posebusters}
\end{table}

Table~\ref{tab:posebusters} compares BridgeDock with the AF3-style co-folding
baseline on PoseBusters. Under the default 20-step setting, BridgeDock improves
all aggregate accuracy metrics and reduces mean ligand RMSD, while the median
RMSD remains unchanged. This suggests that the gain is not mainly obtained by
further polishing already easy examples, but by reducing the tail of more
difficult failures. The few-step results show a second benefit: BridgeDock
remains competitive even with only one or five denoising steps, despite using a
strong ODE sampler for the AF3 baseline. A single BridgeDock step already
exceeds both the 5-step and default 20-step AF3 baselines in lDDT and both
ligand RMSD threshold percentages. These results suggest that apo structures provide an
instance-specific structural prior for co-folding: they anchor the global
protein geometry and reduce the conformational search burden, enabling the
pretrained denoiser to focus on apo-to-holo rearrangement and ligand placement
rather than inferring the entire complex from sequence-derived information
alone. Additional qualitative visual comparisons in
\Cref{app:qualitative_posebusters} further illustrate this behavior on selected
PoseBusters examples.

\subsubsection{PDBBind}

\begin{table}[t]
\centering
\small
\caption{Comparison with prior methods on the PDBBind test set.
Methods are grouped into search-based methods, deep learning-based methods, and co-folding methods. 
The values for SMINA, GNINA, DiffDock-Pocket, Re-Dock, and FlexDock are
literature-reported benchmark results, whereas the Protenix baseline and
BridgeDock results are generated by us. The three flexible docking models use
ligand-derived pocket information as detailed in \Cref{app:docking_baselines}.
For these literature-reported methods, only the percentages below the
2\,\AA{} and 5\,\AA{} ligand RMSD thresholds and median RMSD are available;
unreported metrics are marked as --. Higher is better for lDDT and the
threshold percentages, while lower is better for ligand RMSD.}
\setlength{\tabcolsep}{5pt}
\begin{tabular}{@{}lccccc@{}}
\toprule
& \multicolumn{3}{c}{Docking Accuracy $\uparrow$} & \multicolumn{2}{c}{Ligand RMSD (\AA) $\downarrow$} \\
\cmidrule(lr){2-4}\cmidrule(lr){5-6}
Model & lDDT (\%) & \%<2 $\uparrow$ & \%<5 $\uparrow$ & Mean & Median \\
\midrule

\multicolumn{6}{l}{\textit{Search-based methods}} \\
SMINA         & -- & 3.60 & 20.50 & -- & 7.30 \\
GNINA         & -- & 6.60 & 28.00 & -- & 7.20 \\

\midrule
\multicolumn{6}{l}{\textit{Deep learning-based methods}} \\
DiffDock-Pocket & -- & 41.70 & 74.90 & -- & 2.60 \\
Re-Dock         & -- & 42.90 & 76.40 & -- & 2.40 \\
FlexDock        & -- & 39.70 & --    & -- & 2.50 \\

\midrule
\multicolumn{6}{l}{\textit{Co-folding methods}} \\
Protenix baseline & 88.08 & 69.00 & 86.30 & 2.81 & \textbf{0.99} \\
BridgeDock            & \textbf{92.61} & \textbf{71.00} & \textbf{89.40} & \textbf{2.55} & 1.09 \\

\bottomrule
\end{tabular}
\label{tab:pdbbind_v2020}
\end{table}

As shown in Table~\ref{tab:pdbbind_v2020}, the AF3-style co-folding baseline is
already much stronger than search-based and prior flexible docking methods.
This confirms the value of the pretrained co-folding prior, but also highlights
what is missing from the baseline: it still generates from a
sequence-conditioned diffusion process rather than from the provided apo
complex. BridgeDock preserves this prior and adds an apo-to-holo
bridge, improving the below-2\,\AA{} percentage from 69.0\% to 71.0\%, the
below-5\,\AA{} percentage from 86.3\% to 89.4\%, and lDDT from 88.08\% to
92.61\%, while reducing mean ligand RMSD from 2.81\,\AA{} to 2.55\,\AA{}. The
slightly higher median RMSD suggests that the largest benefit again comes from
better robustness on harder complexes rather than from optimizing only the
easiest poses.

\subsection{Ablations}

\subsubsection{Apo-Conditioned Fine-Tuning}
To isolate whether the gains come simply from providing apo coordinates as
auxiliary conditioning information, we compare BridgeDock with a direct
apo-conditioned fine-tuning baseline. This baseline adds a zero-initialized
fully connected layer to encode the apo structure as an additional conditioning
signal for the diffusion module, but it keeps the original sequence-conditioned
generation process rather than introducing an apo-to-holo bridge. We denote this
baseline as AF3-SFT-Apo.

\begin{table}[t]
\centering
\small
\caption{Ablation results on PoseBusters. The table includes both the
apo-conditioned supervised fine-tuning baseline, denoted as AF3-SFT-Apo, and a
bridge variant without AF3 alignment, denoted as
BridgeDock$_{\mathrm{w/o\ align}}$. Higher is better for lDDT and threshold
percentages, while lower is better for ligand RMSD.}
\setlength{\tabcolsep}{3pt}
\begin{tabular}{@{}lcccccccc@{}}
\toprule
& \multicolumn{3}{c}{Settings} & \multicolumn{3}{c}{Docking Accuracy $\uparrow$} & \multicolumn{2}{c}{Ligand RMSD (\AA) $\downarrow$} \\
\cmidrule(lr){2-4}\cmidrule(lr){5-7}\cmidrule(lr){8-9}
Model & \textit{Apo} & \textit{Bridge} & \textit{AF3-prior} & lDDT (\%) & \%<2 $\uparrow$ & \%<5 $\uparrow$ & Mean & Median \\
\midrule
AF3 baseline & -- & -- & \checkmark & 84.79 & 68.61 & 84.69 & 2.64 & 1.09 \\
AF3-SFT-Apo & \checkmark & -- & \checkmark & 87.53 & 70.14 & 86.12 & 2.48 & \textbf{1.01} \\
BridgeDock$_{\mathrm{w/o\ align}}$ & \checkmark & \checkmark & -- & 85.23 & 68.42 & 86.60 & 2.77 & 1.22 \\
BridgeDock & \checkmark & \checkmark & \checkmark & \textbf{90.41} & \textbf{73.21} & \textbf{89.47} & \textbf{2.26} & 1.09 \\
\bottomrule
\end{tabular}
\label{tab:ablation_comparison}
\end{table}

Table~\ref{tab:ablation_comparison} separates two ways of using the apo
prior. Providing apo coordinates as auxiliary conditioning information already
improves over the AF3 baseline, indicating that the provided structure carries
useful instance-specific information. However, this conditioning-only strategy
leaves the original noise-to-structure generation path unchanged, so the apo
structure can only act as side information. BridgeDock yields stronger gains
across lDDT, threshold-based pose accuracy, and mean ligand RMSD, even though
AF3-SFT-Apo has a slightly lower median RMSD. This pattern suggests that
the bridge formulation is especially helpful for harder cases: rather than
asking the model to infer the bound complex from sequence-derived features with
apo coordinates appended as auxiliary input, it uses the apo complex to shape the
generative trajectory itself and reduce the conformational search burden.

\subsubsection{Effect of AF3 Alignment}
BridgeDock also relies on the AF3-aligned bridge parameterization introduced in
\Cref{sec:af3_aligned_bridge}. To isolate this component, we compare against a bridge
variant that trains and samples directly on the raw bridge states without the
AF3-aligned reparameterization. This control keeps the apo-to-holo bridge
trajectory but removes the mapping that presents each bridge state to the denoiser
as an AF3-style noisy target.

Table~\ref{tab:ablation_comparison} shows that introducing an apo-to-holo bridge
alone is not enough. The unaligned bridge introduces the desired trajectory,
but its raw states are no longer presented in the denoising form on
which the AF3-style backbone was pretrained. As a result, its gains are small
and unstable: the below-5\,\AA{} percentage improves, but the stricter
below-2\,\AA{} percentage and ligand RMSD do not. The AF3-aligned bridge removes
this conflict by keeping the bridge process while mapping each intermediate
state back to an AF3-compatible denoising problem. The large gap between the
unaligned bridge and BridgeDock therefore indicates that the benefit comes from
combining apo-conditioned bridge generation with preservation of the pretrained
co-folding prior, rather than from the bridge objective in isolation.

\section{Conclusion}

In this work, we introduced BridgeDock, a bridge-based generative framework that adapts pretrained co-folding models for molecular docking from apo structures. By formulating docking as an apo-to-holo diffusion bridge and aligning the bridge parameterization with the original AlphaFold~3 denoising objective, our method is able to leverage strong pretrained generative priors while explicitly incorporating structural information from the docking input. Across benchmark evaluations, BridgeDock improves docking accuracy over strong co-folding baselines, and its apo-conditioned initialization further enables efficient generation with fewer denoising steps. These results suggest that bridge-based generative modeling is a promising paradigm for structure-aware docking and, more generally, for transferring large diffusion models from noise-conditioned generation to data-conditioned scientific prediction tasks.

\FloatBarrier
\begin{ack}
This work is supported by the Zhongguancun Academy under Grant C20250601.
We thank the anonymous reviewers for their careful evaluation and constructive feedback, which helped improve the clarity and quality of this work.
\end{ack}

\bibliographystyle{plainnat}
\bibliography{example_paper}

\appendix
\section{AF3 Sampling and Hyperparameter Details}
\label{app:af3_sampling}

At inference, AF3 defines a discrete noise schedule
$\{c_0,c_1,\dots,c_T\}$ according to
\begin{equation}
c_{\tau}
=
\sigma_{\mathrm{data}}
\left(
s_{\max}^{1/p}
+
\frac{\tau}{T}
\left(
s_{\min}^{1/p} - s_{\max}^{1/p}
\right)
\right)^p,
\qquad
\tau = 0,1,\dots,T,
\end{equation}
where $s_{\max}$, $s_{\min}$, $p$, and $T$ are inference hyperparameters.
Given this schedule, the full AF3 reverse sampler is summarized in
\Cref{alg:af3_sampling}.

\begin{algorithm}[H]
\caption{AF3 Inference-Time Sampling}
\label{alg:af3_sampling}
\begin{algorithmic}[1]
\REQUIRE Conditioning features $\mathcal{C}$, noise schedule hyperparameters
$s_{\max}$, $s_{\min}$, $p$, $T$, and sampler hyperparameters
$\gamma_0$, $\gamma_{\min}$, $\lambda$, $\eta$
\STATE Construct the discrete noise schedule $\{c_0,c_1,\dots,c_T\}$.
\STATE Sample $\mathbf{x} \sim c_0 \cdot \mathcal{N}(\mathbf{0}, I)$.
\FOR{$\tau=1$ to $T$}
    \STATE $\mathbf{x} \leftarrow \mathrm{CentreRandomAugmentation}(\mathbf{x})$.
    \IF{$c_{\tau} > \gamma_{\min}$}
        \STATE $\gamma \leftarrow \gamma_0$.
    \ELSE
        \STATE $\gamma \leftarrow 0$.
    \ENDIF
    \STATE $\hat{t} \leftarrow c_{\tau-1}(1+\gamma)$.
    \STATE Sample $\boldsymbol{\epsilon} \sim \mathcal{N}(\mathbf{0}, I)$.
    \STATE $\boldsymbol{\xi} \leftarrow \lambda \sqrt{\hat{t}^2-c_{\tau-1}^2}\,\boldsymbol{\epsilon}$.
    \STATE $\mathbf{x}^{\mathrm{noisy}} \leftarrow \mathbf{x} + \boldsymbol{\xi}$.
    \STATE $\hat{\mathbf{x}}_0 \leftarrow D_\theta(\mathbf{x}^{\mathrm{noisy}}, \hat{t}, \mathcal{C})$.
    \STATE $\boldsymbol{\delta} \leftarrow (\mathbf{x}-\hat{\mathbf{x}}_0)/\hat{t}$.
    \STATE $dt \leftarrow c_{\tau}-\hat{t}$.
    \STATE $\mathbf{x} \leftarrow \mathbf{x}^{\mathrm{noisy}} + \eta\,dt\,\boldsymbol{\delta}$.
\ENDFOR
\STATE \textbf{return} $\mathbf{x}$
\end{algorithmic}
\end{algorithm}

\paragraph{AF3 hyperparameters.}
In \Cref{sec:af3_diffusion}, we use $\sigma_{\mathrm{data}}=16$ for
training-time noise scaling. The training-time noise level is sampled as
\begin{equation}
\hat{t}
=
\sigma_{\mathrm{data}} \cdot \exp\!\left(-1.2 + 1.5 z\right),
\qquad
z \sim \mathcal{N}(0,1),
\end{equation}
which corresponds to a log-normal noise distribution. Before adding noise, AF3
also applies a random rigid augmentation to the ground-truth coordinates; this
step is omitted from the main-text equations for simplicity. For the
inference-time schedule, we set $s_{\max}=160$, $s_{\min}=4\times 10^{-4}$,
$p=7$, and $T=200$. In the AF3 sampler, the additional hyperparameters are
$\gamma_0=0.8$, $\gamma_{\min}=1.0$, $\lambda=1.003$, and $\eta=1.5$.

\FloatBarrier
\section{VE Bridge Schedule and DBIM Inference Details}
\label{app:bridge_inference_details}

\paragraph{DBIM time-step schedule.}
Following the DBIM sampler~\cite{zheng2024diffusion}, we use a decreasing
uniform grid over the VE bridge interval:
\begin{equation}
\tau_n
=
(\sigma_{\max}-10^{-3})
-
\frac{n}{N}
\bigl((\sigma_{\max}-10^{-3})-\sigma_{\min}\bigr),
\qquad
n=0,1,\dots,N.
\label{eq:bridge_schedule}
\end{equation}
Thus $\tau_0=\sigma_{\max}-10^{-3}$ and $\tau_N=\sigma_{\min}$. For any bridge
time $u$, we write $a_u$, $b_u$, and $c_u$ for the VE bridge coefficients in
\Cref{eq:bridge_abc} evaluated at $u$. This shifted start avoids evaluating the
bridge update exactly at the singular endpoint while remaining effectively
apo-conditioned.

\paragraph{Inference-time bridge sampling.}
Given the apo starting structure $\mathbf{x}^{\mathrm{apo}}$, we first center it
and obtain a bootstrap estimate of the clean target structure at noise level
$\sigma_{\max}$:
\begin{equation}
\bar{\mathbf{x}}^{\mathrm{apo}}
=
\mathbf{x}^{\mathrm{apo}}
-
\operatorname{mean}_{\mathrm{atom}}(\mathbf{x}^{\mathrm{apo}}),
\qquad
\hat{\mathbf{x}}_{0}^{(0)}
=
D_{\theta}\!\left(
\bar{\mathbf{x}}^{\mathrm{apo}},\,
\sigma_{\max},\,
\mathcal{C}
\right).
\end{equation}
We then sample the initial bridge state at the first schedule point $\tau_0$:
\begin{equation}
\mathbf{x}_{\tau_0}
=
a_{\tau_0}\mathbf{x}^{\mathrm{apo}}
+
b_{\tau_0}\hat{\mathbf{x}}_{0}^{(0)}
+
c_{\tau_0}\boldsymbol{\epsilon},
\qquad
\boldsymbol{\epsilon} \sim \mathcal{N}(\mathbf{0},I).
\label{eq:bridge_init_sample}
\end{equation}

For each pair of time steps $s=\tau_n$ and $t=\tau_{n+1}$,
$n=0,\dots,N-1$, the current bridge state is centered before denoising:
\begin{equation}
\bar{\mathbf{x}}_s
=
\mathbf{x}_s
-
\operatorname{mean}_{\mathrm{atom}}(\mathbf{x}_s),
\qquad
\hat{\mathbf{x}}_{0}^{(n)}
=
D_{\theta}\!\left(
\bar{\mathbf{x}}_{s},\,
s,\,
\mathcal{C}
\right),
\end{equation}
and perform a DBIM reverse bridge update of the form
\begin{equation}
\mathbf{x}_{t}
=
\alpha_{s\to t}\,\hat{\mathbf{x}}_{0}^{(n)}
+
\beta_{s\to t}\,\mathbf{x}^{\mathrm{apo}}
+
\gamma_{s\to t}\,\mathbf{x}_{s}
+
\mathbb{I}[n<N-1]\,\omega_{s\to t}\,\boldsymbol{\epsilon},
\qquad
\boldsymbol{\epsilon} \sim \mathcal{N}(\mathbf{0},I),
\label{eq:bridge_reverse_update}
\end{equation}
where the coefficients are chosen so that the reverse update is consistent with
the forward VE bridge. In the implementation, these coefficients are written as
\begin{equation}
\tilde{c}_{s\to t}
=
\frac{\sqrt{c_t^2-\omega_{s\to t}^2}}{c_s},
\label{eq:bridge_tildec}
\end{equation}
\begin{equation}
\alpha_{s\to t}=b_t-\tilde{c}_{s\to t} b_s,
\qquad
\beta_{s\to t}=a_t-\tilde{c}_{s\to t} a_s,
\qquad
\gamma_{s\to t}=\tilde{c}_{s\to t},
\label{eq:bridge_reverse_coeffs}
\end{equation}
with stochasticity
\begin{equation}
\omega_{s\to t}
=
\eta\,\alpha_t \rho_t
\sqrt{1-\frac{\rho_t^2}{\rho_s^2}}.
\label{eq:bridge_stochasticity}
\end{equation}
Here, $\eta$ controls the sampling stochasticity and is treated as an
inference hyperparameter; its setting is summarized in
\Cref{app:hyperparameters}. For the VE bridge considered here, we have
$\alpha_t=1$, $\rho_t=t$, and $\rho_s=s$, so the above update reduces to a
VE-specific DBIM reverse sampler from the apo starting structure to the holo
target. Note that the final step does not set $\omega_{s\to t}$ to zero before
computing $\tilde c_{s\to t}$; it only removes the explicit noise-injection term
through the indicator in \Cref{eq:bridge_reverse_update}. After the final step,
$\mathbf{x}_{\tau_N}$ is returned as the predicted all-atom structure. The
complete sampler is summarized in \Cref{alg:dbim_ve_apo}.

\begin{algorithm}[H]
\caption{DBIM Inference with a VE Bridge from Apo Structures}
\label{alg:dbim_ve_apo}
\begin{algorithmic}[1]
\REQUIRE Apo starting structure $\mathbf{x}^{\mathrm{apo}}$, conditioning features $\mathcal{C}$,
number of reverse steps $N$, $\sigma_{\max}$, $\sigma_{\min}$, $\eta$
\STATE Construct the time schedule according to \Cref{eq:bridge_schedule}.
\STATE Compute the VE bridge coefficients according to \Cref{eq:bridge_abc}.
\STATE Use the preconditioned denoiser defined in \Cref{eq:bridge_denoiser}.
\STATE Center the apo structure and bootstrap the first clean-target estimate:
\[
\bar{\mathbf{x}}^{\mathrm{apo}}
\leftarrow
\mathbf{x}^{\mathrm{apo}}
-
\operatorname{mean}_{\mathrm{atom}}(\mathbf{x}^{\mathrm{apo}}),
\qquad
\hat{\mathbf{x}}_0^{(0)}
\leftarrow
D_\theta(\bar{\mathbf{x}}^{\mathrm{apo}},\sigma_{\max},\mathcal{C})
\]
\STATE Sample the initial bridge state according to
\Cref{eq:bridge_init_sample}.
\STATE $\mathbf{x} \leftarrow \mathbf{x}_{\tau_0}$
\FOR{$n=0$ to $N-1$}
    \STATE $s \leftarrow \tau_n,\quad t \leftarrow \tau_{n+1}$
    \STATE Center the current bridge state:
    \[
    \bar{\mathbf{x}}
    \leftarrow
    \mathbf{x}
    -
    \operatorname{mean}_{\mathrm{atom}}(\mathbf{x})
    \]
    \STATE Predict the clean endpoint:
    \[
    \hat{\mathbf{x}}_0 \leftarrow D_\theta(\bar{\mathbf{x}},s,\mathcal{C})
    \]
    \STATE Compute $\omega_{s\to t}$ according to
    \Cref{eq:bridge_stochasticity}.
    \STATE Compute $\tilde c_{s\to t}$, $\alpha_{s\to t}$,
    $\beta_{s\to t}$, and $\gamma_{s\to t}$ according to
    \Cref{eq:bridge_tildec,eq:bridge_reverse_coeffs}.
    \STATE Sample $\boldsymbol{\epsilon} \sim \mathcal N(\mathbf{0},I)$
    \STATE Update the bridge state:
    \[
    \mathbf{x}
    \leftarrow
    \alpha_{s\to t}\hat{\mathbf{x}}_0
    +
    \beta_{s\to t}\mathbf{x}^{\mathrm{apo}}
    +
    \gamma_{s\to t}\mathbf{x}
    +
    \mathbb{I}[n<N-1]\,\omega_{s\to t}\boldsymbol{\epsilon}
    \]
\ENDFOR
\STATE \textbf{return} $\mathbf{x}$
\end{algorithmic}
\end{algorithm}

\FloatBarrier
\section{AF3-Aligned Bridge Algorithms}
\label{app:af3_aligned_algorithms}

\paragraph{Training with AF3-aligned bridge samples.}
The training procedure is summarized in
\Cref{alg:af3_aligned_bridge_train}. The key idea is to first sample a bridge
state between the apo starting structure and the target holo structure, and then
convert that bridge state into an AF3-style noisy target before applying the
denoiser. The time variable and loss weighting remain bridge-based throughout
training. The algorithm makes AF3's random rigid augmentation explicit and
denotes the augmented target by $\tilde{\mathbf{x}}^{\mathrm{GT}}$.

\begin{algorithm}[H]
\caption{Training with AF3-Aligned VE Bridge Samples}
\label{alg:af3_aligned_bridge_train}
\begin{algorithmic}[1]
\REQUIRE Apo starting structure $\mathbf{x}^{\mathrm{apo}}$, ground-truth target $\mathbf{x}^{\mathrm{GT}}$,
conditioning features $\mathcal{C}$, bridge interval
$[\sigma_{\min},\sigma_{\max}]$
\STATE Sample an interior bridge time
\[
t \sim \mathcal{U}(\sigma_{\min},\sigma_{\max})
\]
\STATE Apply random rigid augmentation to the target:
\[
\tilde{\mathbf{x}}^{\mathrm{GT}} \leftarrow \mathrm{Augment}(\mathbf{x}^{\mathrm{GT}})
\]
\STATE Compute VE bridge coefficients
\[
a_t = \frac{t^2}{\sigma_{\max}^2},\qquad
b_t = 1 - \frac{t^2}{\sigma_{\max}^2},\qquad
c_t = t\sqrt{1-\frac{t^2}{\sigma_{\max}^2}}
\]
\STATE Sample Gaussian noise $\boldsymbol{\epsilon} \sim \mathcal{N}(\mathbf{0},I)$
\STATE Sample a bridge state
\[
\mathbf{x}_t \leftarrow a_t \mathbf{x}^{\mathrm{apo}} + b_t \tilde{\mathbf{x}}^{\mathrm{GT}} + c_t \boldsymbol{\epsilon}
\]
\STATE Reparameterize the bridge state into AF3-aligned form
\[
\mathbf{x}_t^{\mathrm{align}} \leftarrow \frac{\mathbf{x}_t - a_t \mathbf{x}^{\mathrm{apo}}}{b_t},
\qquad
\sigma^{\mathrm{align}}_t \leftarrow \frac{c_t}{b_t}
\]
\STATE Predict the clean target endpoint conditioned on the apo structure
\[
\hat{\mathbf{x}}_0 \leftarrow D_\theta(\mathbf{x}_t^{\mathrm{align}}, \sigma^{\mathrm{align}}_t, \mathbf{x}^{\mathrm{apo}}, \mathcal{C})
\]
\STATE Align the ground-truth target to the prediction and compute the
reconstruction loss
\[
\mathcal{L}_{\mathrm{MSE}} \leftarrow \|\hat{\mathbf{x}}_0 - \tilde{\mathbf{x}}^{\mathrm{GT\mbox{-}aligned}}\|^2
\]
\STATE Compute the bridge weight
\[
w_{\mathrm{bridge}}(t) \leftarrow \frac{1}{c_{\mathrm{out}}(t)^2}
\]
\STATE Optimize the bridge objective
\[
\mathcal{L}_{\mathrm{bridge}} \leftarrow w_{\mathrm{bridge}}(t)\,\mathcal{L}_{\mathrm{MSE}}
\]
\STATE \textbf{return} $\mathcal{L}_{\mathrm{bridge}}$
\end{algorithmic}
\end{algorithm}

\paragraph{Inference with AF3-aligned DBIM updates.}
For the AF3-aligned sampler used in our implementation, we use a shifted linear
bridge-time schedule of length $N+1$,
\begin{equation}
\tau_n
=
(\sigma_{\max}-0.1)
-
\frac{n}{N}
\bigl((\sigma_{\max}-0.1)-\sigma_{\min}\bigr),
\qquad
n=0,1,\dots,N.
\label{eq:af3_aligned_bridge_schedule}
\end{equation}
Thus, with $\sigma_{\max}=40.0$ and $\sigma_{\min}=0.4$, the schedule starts at
$\tau_0=39.9$ and ends at $\tau_N=0.4$. Unlike the generic DBIM sampler in
\Cref{alg:dbim_ve_apo}, the AF3-aligned implementation does not bootstrap a
clean endpoint estimate at $\tau_0$; it initializes this estimate to zero and
starts directly from the corresponding near-apo bridge state. Before each
denoising call, the aligned coordinate is also centered atom-wise. The complete
procedure is summarized in \Cref{alg:af3_aligned_bridge_test}.

\begin{algorithm}[H]
\caption{Inference with AF3-Aligned DBIM Reverse Bridge Updates}
\label{alg:af3_aligned_bridge_test}
\begin{algorithmic}[1]
\REQUIRE Apo starting structure $\mathbf{x}^{\mathrm{apo}}$, conditioning features $\mathcal{C}$,
number of reverse steps $N$, $\sigma_{\max}$, $\sigma_{\min}$, $\eta$
\STATE Construct the shifted time schedule according to
\Cref{eq:af3_aligned_bridge_schedule}.
\STATE Use the VE bridge coefficient functions $a_u$, $b_u$, and $c_u$ from
\Cref{eq:bridge_abc}.
\STATE Set the initial clean-endpoint estimate to zero (no bootstrap):
\[
\hat{\mathbf{x}}_0^{(0)} \leftarrow \mathbf{0}
\]
\STATE Sample the initial near-apo bridge state
\[
\mathbf{x}
\leftarrow
a_{\tau_0}\mathbf{x}^{\mathrm{apo}}
+
b_{\tau_0}\hat{\mathbf{x}}_0^{(0)}
+
c_{\tau_0}\boldsymbol{\epsilon}
=
a_{\tau_0}\mathbf{x}^{\mathrm{apo}}
+
c_{\tau_0}\boldsymbol{\epsilon},
\qquad
\boldsymbol{\epsilon} \sim \mathcal{N}(\mathbf{0},I)
\]
\FOR{$n = 0$ to $N-1$}
    \STATE $s \leftarrow \tau_n,\quad t \leftarrow \tau_{n+1}$
    \STATE Reparameterize the current bridge state into AF3-aligned form
    \[
    \mathbf{x}_s^{\mathrm{align}} \leftarrow \frac{\mathbf{x} - a_s \mathbf{x}^{\mathrm{apo}}}{b_s},
    \qquad
    \sigma^{\mathrm{align}}_s \leftarrow \frac{c_s}{b_s}
\]
    \STATE Center the aligned coordinate before denoising
    \[
    \mathbf{x}_s^{\mathrm{align}}
    \leftarrow
    \mathbf{x}_s^{\mathrm{align}}
    -
    \operatorname{mean}_{\mathrm{atom}}
    \bigl(\mathbf{x}_s^{\mathrm{align}}\bigr)
    \]
    \STATE Predict the clean target endpoint conditioned on the apo structure
    \[
    \hat{\mathbf{x}}_0 \leftarrow D_\theta(\mathbf{x}_s^{\mathrm{align}}, \sigma^{\mathrm{align}}_s, \mathbf{x}^{\mathrm{apo}}, \mathcal{C})
    \]
    \STATE Compute the VE-bridge DBIM stochasticity
    \[
    \omega_{s\to t}
    \leftarrow
    \eta\,t\sqrt{1-\frac{t^2}{s^2}}
    \]
    \STATE Compute $\tilde{c}_{s\to t}$ using the unmasked stochasticity
    \[
    \tilde{c}_{s\to t}
    \leftarrow
    \frac{\sqrt{c_t^2-\omega_{s\to t}^2}}{c_s}
    \]
    \STATE Sample $\boldsymbol{\epsilon} \sim \mathcal{N}(\mathbf{0},I)$
    \STATE Update the bridge state
    \[
    \mathbf{x} \leftarrow
    \bigl(b_t-\tilde{c}_{s\to t}b_s\bigr)\hat{\mathbf{x}}_0
    +
    \bigl(a_t-\tilde{c}_{s\to t}a_s\bigr)\mathbf{x}^{\mathrm{apo}}
    +
    \tilde{c}_{s\to t}\mathbf{x}
    +
    \mathbb{I}[n<N-1]\,\omega_{s\to t}\boldsymbol{\epsilon}
    \]
\ENDFOR
\STATE \textbf{return} $\mathbf{x}$
\end{algorithmic}
\end{algorithm}

\FloatBarrier
\section{Bridge Hyperparameter Settings}
\label{app:hyperparameters}

This appendix collects the bridge-specific hyperparameter values omitted from
\Cref{sec:bridge_diffusion}. Throughout the main text, we use
``hyperparameter'' to refer to fixed user-specified constants that determine
the training or sampling procedure but are not predicted by the model.

\paragraph{Bridge hyperparameters.}
For the VE bridge in \Cref{sec:bridge_diffusion}, we set
$\sigma_{\min}=0.4$ and $\sigma_{\max}=40.0$. For bridge preconditioning, we
use $\sigma_{\mathrm{data}}=16$, $\sigma_{\mathrm{data,end}}=16$,
$\mathrm{cov}_{xy}=0$, and $\kappa=1$. For inference, we use a uniform bridge
time schedule with the same $\sigma_{\min}$ and $\sigma_{\max}$; in the
AF3-aligned implementation, the first inference time is shifted to
$\sigma_{\max}-0.1=39.9$ as described in
\Cref{eq:af3_aligned_bridge_schedule}. The remaining inference hyperparameters,
including the reverse-step count $N$ and stochasticity coefficient $\eta$, are
implementation-level tuning parameters.

\subsection{Sensitivity to the Diffusion-Time Offset}
\label{app:offset_sensitivity}

The offset is required to avoid the singular endpoint of the AF3-aligned bridge
parameterization. At $t=\sigma_{\max}$, we have $b_t=0$; consequently, both
$\mathbf{x}^{\mathrm{align}}_t=(\mathbf{x}_t-a_t\mathbf{x}^{\mathrm{apo}})/b_t$
and $\sigma^{\mathrm{align}}_t=c_t/b_t$ are undefined. Starting from
$\sigma_{\max}-\delta$ avoids this division by zero and improves numerical
conditioning while keeping the initial state close to the apo endpoint.

We use the schedule from $\tau_0=\sigma_{\max}-\delta$ to
$\tau_N=\sigma_{\min}$, as described in
\Cref{eq:af3_aligned_bridge_schedule} for the default offset $\delta=0.1$.
Here, $\sigma_{\max}=40$ and $\sigma_{\min}=0.4$ are the bridge-time boundaries
used during training. The end point is kept at $\sigma_{\min}$ to avoid
extrapolating beyond the trained time range.

We additionally evaluated the sensitivity to $\delta$ on PoseBusters using
20 sampling steps, as shown in \Cref{tab:offset_sensitivity}.

\begin{table}[H]
\centering
\small
\caption{Sensitivity to the diffusion-time offset on PoseBusters using
20 sampling steps. The start time is $\tau_0=\sigma_{\max}-\delta$.
Higher is better for lDDT and the ligand RMSD threshold percentages; lower is
better for mean and median ligand RMSD.}
\label{tab:offset_sensitivity}
\setlength{\tabcolsep}{6pt}
\begin{tabular}{@{}lccccc@{}}
\toprule
& \multicolumn{3}{c}{Docking Accuracy $\uparrow$} & \multicolumn{2}{c}{Ligand RMSD (\AA) $\downarrow$} \\
\cmidrule(lr){2-4}\cmidrule(lr){5-6}
Offset $\delta$ ($\tau_0$) & lDDT (\%) & \%$<2$\,\AA{} & \%$<5$\,\AA{} & Mean & Median \\
\midrule
1 (39.0)       & 90.01 & 71.77 & 89.47 & 2.26 & 1.07 \\
0.1 (39.9)     & 90.41 & 73.21 & 89.47 & 2.26 & 1.09 \\
0.01 (39.99)   & 90.46 & 72.25 & 89.95 & 2.18 & 1.06 \\
0.001 (39.999) & 90.43 & 72.73 & 90.43 & 2.16 & 1.07 \\
\bottomrule
\end{tabular}
\end{table}

The results are stable over three orders of magnitude, indicating that BridgeDock
is not sensitive to the precise offset. We use $\delta=0.1$ because it provides
a safe numerical margin from the singular endpoint while achieving the highest
success rate below 2\,\AA{} and near-best overall performance.

\FloatBarrier
\section{Pocket-Aware Settings of Flexible Docking Baselines}
\label{app:docking_baselines}

The results for DiffDock-Pocket~\cite{plainer2023diffdock},
Re-Dock~\cite{huang2024re}, and FlexDock~\cite{corsocomposing} are taken from
their original publications, which evaluate these methods using pocket-aware
protein inputs. DiffDock-Pocket defines the pocket center as the mean position
of the apo-protein $C_{\alpha}$ atoms belonging to residues that contain at least
one atom within 5\,\AA{} of the ground-truth ligand. It retains residues within
a pocket radius defined by the ligand circumradius plus a 10\,\AA{} context
buffer, and treats residues containing an atom within 3.5\,\AA{} of the ligand
as flexible. Re-Dock defines the pocket as all protein residues within
8\,\AA{} of the ligand and constructs its pocket-ligand heterogeneous graph
from the resulting cropped pocket, using the corresponding apo-pocket structure
as receptor input in the apo-docking evaluation. FlexDock defines pocket
residues as those containing at least one heavy atom within 5\,\AA{} of any
ground-truth ligand atom. Its pocket center is computed from the corresponding
$C_{\alpha}$ coordinates in the apo structure, and the model additionally
includes neighboring residues whose apo $C_{\alpha}$ atoms lie within
20\,\AA{} of this center. Thus, all three baselines receive ligand-derived
pocket information and operate on a cropped protein region. In contrast,
BridgeDock receives the complete apo-protein structure, is not provided with the
crystallographic ligand position, pocket center, or pocket-residue annotations,
and initializes the ligand at the geometric center of the full apo protein.

\FloatBarrier
\section{Additional Evaluation on FoldBench}
\label{app:foldbench}

We additionally evaluate BridgeDock with both the Protenix-Mini and
standard-sized Protenix-Base v1.0.0 backbones on the protein-ligand subset of
FoldBench~\cite{xu2026foldbench}. The standard backbone uses the released
\texttt{protenix\_base\_default\_v1.0.0} checkpoint, and we refer to its
BridgeDock variant as BridgeDock-v1. We retain 509 of the 558 targets after
applying a maximum input length of 1,000 tokens, excluding the 49 longer targets
before inference using the same criterion for all methods. For each target, we
generate five predictions using a single random seed and select the prediction
with the highest model confidence. The Protenix-Base v1.0.0 baseline uses its
standard database-retrieved templates, and both models based on this backbone
use 200 diffusion steps. A prediction is considered successful when its
pocket-aligned ligand RMSD is below 2\,\AA{} and its lDDT-PLI exceeds 0.8.

\begin{table}[H]
\centering
\small
\caption{FoldBench protein-ligand results on 509 targets with a single random
seed and five predictions per target. We report confidence-ranked top-1
performance. Higher is better for all metrics; bold indicates the better result
within each backbone setting.}
\label{tab:foldbench_backbones}
\setlength{\tabcolsep}{5pt}
\begin{tabular}{@{}llccc@{}}
\toprule
Backbone & Method & Success rate (\%) $\uparrow$ & lDDT-LP $\uparrow$ & lDDT-PLI $\uparrow$ \\
\midrule
Protenix-Mini & Baseline & 41.65 & 0.8153 & 0.6145 \\
Protenix-Mini & BridgeDock & \textbf{42.63} & \textbf{0.8829} & \textbf{0.6377} \\
\midrule
Protenix-Base v1.0.0 & Baseline & 56.97 & 0.9107 & 0.7251 \\
Protenix-Base v1.0.0 & BridgeDock-v1 & \textbf{59.33} & \textbf{0.9112} & \textbf{0.7504} \\
\bottomrule
\end{tabular}
\end{table}

As shown in \Cref{tab:foldbench_backbones}, BridgeDock improves the success
rate from 41.65\% to 42.63\% with Protenix-Mini and from 56.97\% to 59.33\%
with Protenix-Base v1.0.0, corresponding to gains of 0.98 and 2.36 percentage
points, respectively. Both backbone settings also show improvements in lDDT-LP
and lDDT-PLI, indicating better pocket and protein-ligand interface accuracy.

\FloatBarrier
\section{Qualitative Results on PoseBusters}
\label{app:qualitative_posebusters}

\begin{figure}[p]
    \centering
    \includegraphics[width=\linewidth, trim=212bp 22bp 245bp 9bp, clip]{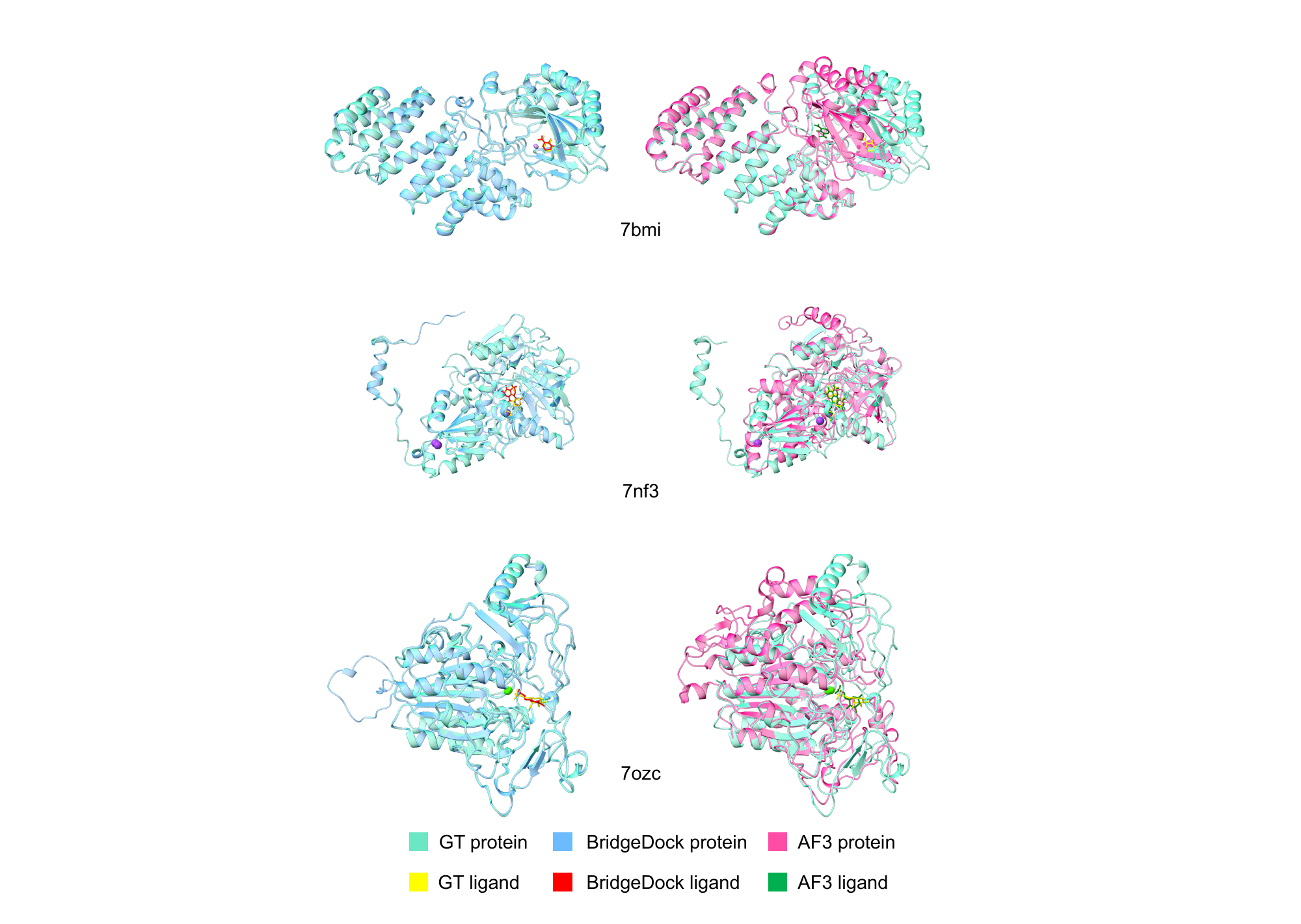}
    \caption{
    Qualitative structural overlays for PoseBusters targets 7bmi (top),
    7nf3 (middle), and 7ozc (bottom). Each row compares BridgeDock (left)
    and the AF3-style Protenix baseline (right) against the same ground-truth
    holo complex. Ground-truth proteins and ligands are shown in cyan and
    yellow, respectively; BridgeDock predictions are shown in blue and red;
    and Protenix predictions are shown in magenta and green. Closer overlap
    of the protein cartoons indicates better agreement in receptor
    conformation, while closer overlap of the ligand sticks indicates better
    agreement in ligand pose. In these examples, BridgeDock more closely
    matches both the ground-truth receptor and ligand, whereas the Protenix
    baseline shows larger deviations in receptor conformation and/or ligand
    placement.
    }
    \label{fig:posebusters_examples_1}
\end{figure}

\begin{figure}[p]
    \centering
    \includegraphics[width=\linewidth, trim=212bp 22bp 245bp 9bp, clip]{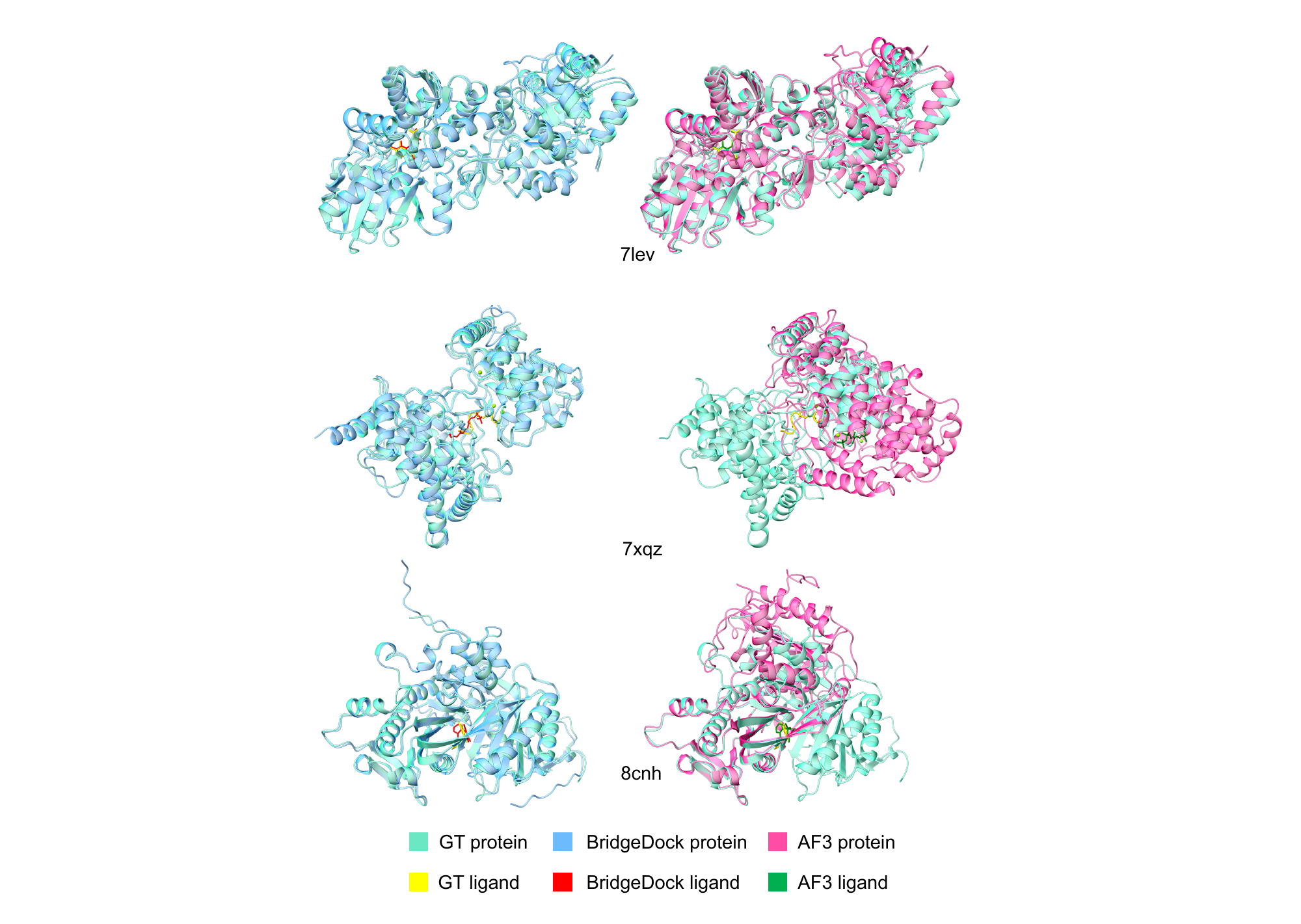}
    \caption{
    Additional qualitative structural overlays for PoseBusters targets 7lev
    (top), 7xqz (middle), and 8cnh (bottom). Each row compares BridgeDock
    (left) and the AF3-style Protenix baseline (right) against the same
    ground-truth holo complex. Ground-truth proteins and ligands are shown
    in cyan and yellow, respectively; BridgeDock predictions are shown in
    blue and red; and Protenix predictions are shown in magenta and green.
    The protein and ligand overlays visualize agreement in receptor
    conformation and ligand pose, respectively. Across the shown examples,
    BridgeDock exhibits closer joint agreement with the ground-truth
    protein-ligand complexes, particularly around the protein-ligand
    interface.
    }
    \label{fig:posebusters_examples_2}
\end{figure}

To complement the aggregate PoseBusters results in \Cref{tab:posebusters}, we
provide qualitative comparisons on six selected PoseBusters examples in
\Cref{fig:posebusters_examples_1,fig:posebusters_examples_2}. Each row shows
the same target under two structural overlays: the left panel compares
BridgeDock with the ground-truth holo complex, while the right panel compares
the AF3 baseline with the same ground-truth structure. The protein and
ligand of the ground-truth complex are shown in cyan and yellow, respectively;
BridgeDock predictions are shown in blue and red; and AF3 baseline predictions
are shown in magenta and green.

Across these examples, BridgeDock generally predicts protein structures that
align more closely with the ground-truth holo conformations, while also placing
the ligand closer to the ground-truth binding pose. The
AF3 baseline often recovers a broadly plausible complex, but the overlays
show larger deviations around the binding pocket and, in several cases, more
visible ligand displacement relative to the ground-truth ligand. These visual
examples are consistent with the quantitative trend in \Cref{tab:posebusters}:
starting the generative process from an apo-informed bridge state helps reduce
large docking errors and improves the fraction of poses within strict ligand
RMSD thresholds. The comparison also illustrates why the gain is not limited to
protein backbone alignment; the improvement is most apparent near the
protein-ligand interface, where accurate pocket geometry and ligand placement
must be resolved jointly.

\FloatBarrier
\section{Limitations and Broader Impact}
\label{app:limitations_impact}

\paragraph{Limitations.}
BridgeDock relies on the quality of the provided apo protein structure and the
initial ligand conformation. Large apo-structure errors, missing cofactors or
modified residues, or pocket states that are poorly represented in the training
data may reduce docking accuracy. Our evaluation focuses on PoseBusters v2 and
a time-based PDBBind v2020 split; broader prospective validation and
experimental follow-up are needed before using predictions as decisive evidence
in downstream discovery. The method also inherits the cost of fine-tuning an
AF3-style backbone: although the bridge formulation improves few-step sampling,
training still requires multi-GPU resources.

\paragraph{Broader impact.}
More accurate and efficient protein-ligand docking could help prioritize
experiments in structure-based drug discovery and improve mechanistic
understanding of molecular interactions. At the same time, incorrect predictions
could mislead downstream experimental decisions if used without expert review,
and improved molecular modeling may in principle contribute to dual-use
chemistry workflows. BridgeDock is intended as a research tool; practical use
should include uncertainty-aware filtering, domain-expert inspection, and
experimental validation.

\FloatBarrier
\newpage
\section*{NeurIPS Paper Checklist}

\begin{enumerate}

\item {\bf Claims}
    \item[] Question: Do the main claims made in the abstract and introduction accurately reflect the paper's contributions and scope?
    \item[] Answer: \answerYes{}.
    \item[] Justification: The abstract and introduction state the BridgeDock formulation, its reliance on apo-conditioned docking inputs, and the main empirical claims. These claims are evaluated in the main experiments on PoseBusters and PDBBind and in the ablation study.
    \item[] Guidelines:
    \begin{itemize}
        \item The answer \answerNA{} means that the abstract and introduction do not include the claims made in the paper.
        \item The abstract and/or introduction should clearly state the claims made, including the contributions made in the paper and important assumptions and limitations. A \answerNo{} or \answerNA{} answer to this question will not be perceived well by the reviewers. 
        \item The claims made should match theoretical and experimental results, and reflect how much the results can be expected to generalize to other settings. 
        \item It is fine to include aspirational goals as motivation as long as it is clear that these goals are not attained by the paper. 
    \end{itemize}

\item {\bf Limitations}
    \item[] Question: Does the paper discuss the limitations of the work performed by the authors?
    \item[] Answer: \answerYes{}.
    \item[] Justification: Limitations are discussed in \Cref{app:limitations_impact}, including dependence on apo-structure quality, benchmark scope, and multi-GPU training cost.
    \item[] Guidelines:
    \begin{itemize}
        \item The answer \answerNA{} means that the paper has no limitation while the answer \answerNo{} means that the paper has limitations, but those are not discussed in the paper. 
        \item The authors are encouraged to create a separate ``Limitations'' section in their paper.
        \item The paper should point out any strong assumptions and how robust the results are to violations of these assumptions (e.g., independence assumptions, noiseless settings, model well-specification, asymptotic approximations only holding locally). The authors should reflect on how these assumptions might be violated in practice and what the implications would be.
        \item The authors should reflect on the scope of the claims made, e.g., if the approach was only tested on a few datasets or with a few runs. In general, empirical results often depend on implicit assumptions, which should be articulated.
        \item The authors should reflect on the factors that influence the performance of the approach. For example, a facial recognition algorithm may perform poorly when image resolution is low or images are taken in low lighting. Or a speech-to-text system might not be used reliably to provide closed captions for online lectures because it fails to handle technical jargon.
        \item The authors should discuss the computational efficiency of the proposed algorithms and how they scale with dataset size.
        \item If applicable, the authors should discuss possible limitations of their approach to address problems of privacy and fairness.
        \item While the authors might fear that complete honesty about limitations might be used by reviewers as grounds for rejection, a worse outcome might be that reviewers discover limitations that aren't acknowledged in the paper. The authors should use their best judgment and recognize that individual actions in favor of transparency play an important role in developing norms that preserve the integrity of the community. Reviewers will be specifically instructed to not penalize honesty concerning limitations.
    \end{itemize}

\item {\bf Theory assumptions and proofs}
    \item[] Question: For each theoretical result, does the paper provide the full set of assumptions and a complete (and correct) proof?
    \item[] Answer: \answerNA{}.
    \item[] Justification: The paper does not present formal theoretical theorems or proof-based claims; the mathematical content defines the diffusion bridge formulation and sampling/training algorithms.
    \item[] Guidelines:
    \begin{itemize}
        \item The answer \answerNA{} means that the paper does not include theoretical results. 
        \item All the theorems, formulas, and proofs in the paper should be numbered and cross-referenced.
        \item All assumptions should be clearly stated or referenced in the statement of any theorems.
        \item The proofs can either appear in the main paper or the supplemental material, but if they appear in the supplemental material, the authors are encouraged to provide a short proof sketch to provide intuition. 
        \item Inversely, any informal proof provided in the core of the paper should be complemented by formal proofs provided in appendix or supplemental material.
        \item Theorems and Lemmas that the proof relies upon should be properly referenced. 
    \end{itemize}

    \item {\bf Experimental result reproducibility}
    \item[] Question: Does the paper fully disclose all the information needed to reproduce the main experimental results of the paper to the extent that it affects the main claims and/or conclusions of the paper (regardless of whether the code and data are provided or not)?
    \item[] Answer: \answerYes{}.
    \item[] Justification: The method, objective, sampling procedures, datasets, baselines, metrics, and bridge hyperparameters are described in the main text and appendices. The main experiments also specify the training data scale, training steps, and compute used for fine-tuning.
    \item[] Guidelines:
    \begin{itemize}
        \item The answer \answerNA{} means that the paper does not include experiments.
        \item If the paper includes experiments, a \answerNo{} answer to this question will not be perceived well by the reviewers: Making the paper reproducible is important, regardless of whether the code and data are provided or not.
        \item If the contribution is a dataset and\slash or model, the authors should describe the steps taken to make their results reproducible or verifiable. 
        \item Depending on the contribution, reproducibility can be accomplished in various ways. For example, if the contribution is a novel architecture, describing the architecture fully might suffice, or if the contribution is a specific model and empirical evaluation, it may be necessary to either make it possible for others to replicate the model with the same dataset, or provide access to the model. In general. releasing code and data is often one good way to accomplish this, but reproducibility can also be provided via detailed instructions for how to replicate the results, access to a hosted model (e.g., in the case of a large language model), releasing of a model checkpoint, or other means that are appropriate to the research performed.
        \item While NeurIPS does not require releasing code, the conference does require all submissions to provide some reasonable avenue for reproducibility, which may depend on the nature of the contribution. For example
        \begin{enumerate}
            \item If the contribution is primarily a new algorithm, the paper should make it clear how to reproduce that algorithm.
            \item If the contribution is primarily a new model architecture, the paper should describe the architecture clearly and fully.
            \item If the contribution is a new model (e.g., a large language model), then there should either be a way to access this model for reproducing the results or a way to reproduce the model (e.g., with an open-source dataset or instructions for how to construct the dataset).
            \item We recognize that reproducibility may be tricky in some cases, in which case authors are welcome to describe the particular way they provide for reproducibility. In the case of closed-source models, it may be that access to the model is limited in some way (e.g., to registered users), but it should be possible for other researchers to have some path to reproducing or verifying the results.
        \end{enumerate}
    \end{itemize}

\item {\bf Open access to data and code}
    \item[] Question: Does the paper provide open access to the data and code, with sufficient instructions to faithfully reproduce the main experimental results, as described in supplemental material?
    \item[] Answer: \answerNo{}.
    \item[] Justification: The experiments use public benchmark datasets and an open-source AF3-style implementation, but this submission does not yet include a complete anonymized code release, checkpoints, or executable reproduction scripts for all reported results.
    \item[] Guidelines:
    \begin{itemize}
        \item The answer \answerNA{} means that paper does not include experiments requiring code.
        \item Please see the NeurIPS code and data submission guidelines (\url{https://neurips.cc/public/guides/CodeSubmissionPolicy}) for more details.
        \item While we encourage the release of code and data, we understand that this might not be possible, so \answerNo{} is an acceptable answer. Papers cannot be rejected simply for not including code, unless this is central to the contribution (e.g., for a new open-source benchmark).
        \item The instructions should contain the exact command and environment needed to run to reproduce the results. See the NeurIPS code and data submission guidelines (\url{https://neurips.cc/public/guides/CodeSubmissionPolicy}) for more details.
        \item The authors should provide instructions on data access and preparation, including how to access the raw data, preprocessed data, intermediate data, and generated data, etc.
        \item The authors should provide scripts to reproduce all experimental results for the new proposed method and baselines. If only a subset of experiments are reproducible, they should state which ones are omitted from the script and why.
        \item At submission time, to preserve anonymity, the authors should release anonymized versions (if applicable).
        \item Providing as much information as possible in supplemental material (appended to the paper) is recommended, but including URLs to data and code is permitted.
    \end{itemize}

\item {\bf Experimental setting/details}
    \item[] Question: Does the paper specify all the training and test details (e.g., data splits, hyperparameters, how they were chosen, type of optimizer) necessary to understand the results?
    \item[] Answer: \answerYes{}.
    \item[] Justification: The experimental setup describes the training corpus size, training steps, datasets, apo input construction, evaluation metrics, baselines, and denoising budgets. Additional algorithmic and hyperparameter details are provided in the appendices.
    \item[] Guidelines:
    \begin{itemize}
        \item The answer \answerNA{} means that the paper does not include experiments.
        \item The experimental setting should be presented in the core of the paper to a level of detail that is necessary to appreciate the results and make sense of them.
        \item The full details can be provided either with the code, in appendix, or as supplemental material.
    \end{itemize}

\item {\bf Experiment statistical significance}
    \item[] Question: Does the paper report error bars suitably and correctly defined or other appropriate information about the statistical significance of the experiments?
    \item[] Answer: \answerNo{}.
    \item[] Justification: The paper reports aggregate metrics on fixed benchmark test sets, but does not include error bars, confidence intervals, or repeated-run statistical significance tests. The reported metrics should therefore be interpreted as benchmark estimates rather than uncertainty-calibrated measurements.
    \item[] Guidelines:
    \begin{itemize}
        \item The answer \answerNA{} means that the paper does not include experiments.
        \item The authors should answer \answerYes{} if the results are accompanied by error bars, confidence intervals, or statistical significance tests, at least for the experiments that support the main claims of the paper.
        \item The factors of variability that the error bars are capturing should be clearly stated (for example, train/test split, initialization, random drawing of some parameter, or overall run with given experimental conditions).
        \item The method for calculating the error bars should be explained (closed form formula, call to a library function, bootstrap, etc.)
        \item The assumptions made should be given (e.g., Normally distributed errors).
        \item It should be clear whether the error bar is the standard deviation or the standard error of the mean.
        \item It is OK to report 1-sigma error bars, but one should state it. The authors should preferably report a 2-sigma error bar than state that they have a 96\% CI, if the hypothesis of Normality of errors is not verified.
        \item For asymmetric distributions, the authors should be careful not to show in tables or figures symmetric error bars that would yield results that are out of range (e.g., negative error rates).
        \item If error bars are reported in tables or plots, the authors should explain in the text how they were calculated and reference the corresponding figures or tables in the text.
    \end{itemize}

\item {\bf Experiments compute resources}
    \item[] Question: For each experiment, does the paper provide sufficient information on the computer resources (type of compute workers, memory, time of execution) needed to reproduce the experiments?
    \item[] Answer: \answerNo{}.
    \item[] Justification: The paper reports the main fine-tuning resource usage as 8$\times$A100 GPUs for approximately five days, but does not exhaustively itemize memory, inference wall-clock time, or total compute for each individual experimental run.
    \item[] Guidelines:
    \begin{itemize}
        \item The answer \answerNA{} means that the paper does not include experiments.
        \item The paper should indicate the type of compute workers CPU or GPU, internal cluster, or cloud provider, including relevant memory and storage.
        \item The paper should provide the amount of compute required for each of the individual experimental runs as well as estimate the total compute. 
        \item The paper should disclose whether the full research project required more compute than the experiments reported in the paper (e.g., preliminary or failed experiments that didn't make it into the paper). 
    \end{itemize}
    
\item {\bf Code of ethics}
    \item[] Question: Does the research conducted in the paper conform, in every respect, with the NeurIPS Code of Ethics \url{https://neurips.cc/public/EthicsGuidelines}?
    \item[] Answer: \answerYes{}.
    \item[] Justification: The work uses public biomolecular structure benchmarks and does not involve human subjects, private data, or deceptive data collection. The intended use is scientific structure prediction and molecular docking research.
    \item[] Guidelines:
    \begin{itemize}
        \item The answer \answerNA{} means that the authors have not reviewed the NeurIPS Code of Ethics.
        \item If the authors answer \answerNo, they should explain the special circumstances that require a deviation from the Code of Ethics.
        \item The authors should make sure to preserve anonymity (e.g., if there is a special consideration due to laws or regulations in their jurisdiction).
    \end{itemize}

\item {\bf Broader impacts}
    \item[] Question: Does the paper discuss both potential positive societal impacts and negative societal impacts of the work performed?
    \item[] Answer: \answerYes{}.
    \item[] Justification: \Cref{app:limitations_impact} discusses potential benefits for structure-based drug discovery and possible risks from incorrect predictions or dual-use molecular modeling workflows.
    \item[] Guidelines:
    \begin{itemize}
        \item The answer \answerNA{} means that there is no societal impact of the work performed.
        \item If the authors answer \answerNA{} or \answerNo, they should explain why their work has no societal impact or why the paper does not address societal impact.
        \item Examples of negative societal impacts include potential malicious or unintended uses (e.g., disinformation, generating fake profiles, surveillance), fairness considerations (e.g., deployment of technologies that could make decisions that unfairly impact specific groups), privacy considerations, and security considerations.
        \item The conference expects that many papers will be foundational research and not tied to particular applications, let alone deployments. However, if there is a direct path to any negative applications, the authors should point it out. For example, it is legitimate to point out that an improvement in the quality of generative models could be used to generate Deepfakes for disinformation. On the other hand, it is not needed to point out that a generic algorithm for optimizing neural networks could enable people to train models that generate Deepfakes faster.
        \item The authors should consider possible harms that could arise when the technology is being used as intended and functioning correctly, harms that could arise when the technology is being used as intended but gives incorrect results, and harms following from (intentional or unintentional) misuse of the technology.
        \item If there are negative societal impacts, the authors could also discuss possible mitigation strategies (e.g., gated release of models, providing defenses in addition to attacks, mechanisms for monitoring misuse, mechanisms to monitor how a system learns from feedback over time, improving the efficiency and accessibility of ML).
    \end{itemize}
    
\item {\bf Safeguards}
    \item[] Question: Does the paper describe safeguards that have been put in place for responsible release of data or models that have a high risk for misuse (e.g., pre-trained language models, image generators, or scraped datasets)?
    \item[] Answer: \answerNA{}.
    \item[] Justification: The paper does not release scraped human data, generative media models, or other assets of the types highlighted as high risk in this checklist item. If model checkpoints are later released, they should be accompanied by appropriate research-use documentation.
    \item[] Guidelines:
    \begin{itemize}
        \item The answer \answerNA{} means that the paper poses no such risks.
        \item Released models that have a high risk for misuse or dual-use should be released with necessary safeguards to allow for controlled use of the model, for example by requiring that users adhere to usage guidelines or restrictions to access the model or implementing safety filters. 
        \item Datasets that have been scraped from the Internet could pose safety risks. The authors should describe how they avoided releasing unsafe images.
        \item We recognize that providing effective safeguards is challenging, and many papers do not require this, but we encourage authors to take this into account and make a best faith effort.
    \end{itemize}

\item {\bf Licenses for existing assets}
    \item[] Question: Are the creators or original owners of assets (e.g., code, data, models), used in the paper, properly credited and are the license and terms of use explicitly mentioned and properly respected?
    \item[] Answer: \answerNo{}.
    \item[] Justification: The paper cites the main datasets, baselines, and the AF3-style implementation used in the study, but the manuscript does not yet explicitly enumerate the licenses and terms of use for all existing assets.
    \item[] Guidelines:
    \begin{itemize}
        \item The answer \answerNA{} means that the paper does not use existing assets.
        \item The authors should cite the original paper that produced the code package or dataset.
        \item The authors should state which version of the asset is used and, if possible, include a URL.
        \item The name of the license (e.g., CC-BY 4.0) should be included for each asset.
        \item For scraped data from a particular source (e.g., website), the copyright and terms of service of that source should be provided.
        \item If assets are released, the license, copyright information, and terms of use in the package should be provided. For popular datasets, \url{paperswithcode.com/datasets} has curated licenses for some datasets. Their licensing guide can help determine the license of a dataset.
        \item For existing datasets that are re-packaged, both the original license and the license of the derived asset (if it has changed) should be provided.
        \item If this information is not available online, the authors are encouraged to reach out to the asset's creators.
    \end{itemize}

\item {\bf New assets}
    \item[] Question: Are new assets introduced in the paper well documented and is the documentation provided alongside the assets?
    \item[] Answer: \answerNA{}.
    \item[] Justification: The paper does not introduce a new dataset or release a new asset package as part of the anonymous submission. If code or checkpoints are released later, documentation and license information should be provided with the release.
    \item[] Guidelines:
    \begin{itemize}
        \item The answer \answerNA{} means that the paper does not release new assets.
        \item Researchers should communicate the details of the dataset\slash code\slash model as part of their submissions via structured templates. This includes details about training, license, limitations, etc. 
        \item The paper should discuss whether and how consent was obtained from people whose asset is used.
        \item At submission time, remember to anonymize your assets (if applicable). You can either create an anonymized URL or include an anonymized zip file.
    \end{itemize}

\item {\bf Crowdsourcing and research with human subjects}
    \item[] Question: For crowdsourcing experiments and research with human subjects, does the paper include the full text of instructions given to participants and screenshots, if applicable, as well as details about compensation (if any)? 
    \item[] Answer: \answerNA{}.
    \item[] Justification: The paper does not involve crowdsourcing experiments or research with human subjects.
    \item[] Guidelines:
    \begin{itemize}
        \item The answer \answerNA{} means that the paper does not involve crowdsourcing nor research with human subjects.
        \item Including this information in the supplemental material is fine, but if the main contribution of the paper involves human subjects, then as much detail as possible should be included in the main paper. 
        \item According to the NeurIPS Code of Ethics, workers involved in data collection, curation, or other labor should be paid at least the minimum wage in the country of the data collector. 
    \end{itemize}

\item {\bf Institutional review board (IRB) approvals or equivalent for research with human subjects}
    \item[] Question: Does the paper describe potential risks incurred by study participants, whether such risks were disclosed to the subjects, and whether Institutional Review Board (IRB) approvals (or an equivalent approval/review based on the requirements of your country or institution) were obtained?
    \item[] Answer: \answerNA{}.
    \item[] Justification: The paper does not involve crowdsourcing experiments or research with human subjects, so IRB approval or equivalent human-subjects review is not applicable.
    \item[] Guidelines:
    \begin{itemize}
        \item The answer \answerNA{} means that the paper does not involve crowdsourcing nor research with human subjects.
        \item Depending on the country in which research is conducted, IRB approval (or equivalent) may be required for any human subjects research. If you obtained IRB approval, you should clearly state this in the paper. 
        \item We recognize that the procedures for this may vary significantly between institutions and locations, and we expect authors to adhere to the NeurIPS Code of Ethics and the guidelines for their institution. 
        \item For initial submissions, do not include any information that would break anonymity (if applicable), such as the institution conducting the review.
    \end{itemize}

\item {\bf Declaration of LLM usage}
    \item[] Question: Does the paper describe the usage of LLMs if it is an important, original, or non-standard component of the core methods in this research? Note that if the LLM is used only for writing, editing, or formatting purposes and does \emph{not} impact the core methodology, scientific rigor, or originality of the research, declaration is not required.
    \item[] Answer: \answerNA{}.
    \item[] Justification: The core method does not use LLMs as an important, original, or non-standard component. Any ordinary writing, editing, or formatting assistance is outside the scope of this disclosure item.
    \item[] Guidelines:
    \begin{itemize}
        \item The answer \answerNA{} means that the core method development in this research does not involve LLMs as any important, original, or non-standard components.
        \item Please refer to our LLM policy in the NeurIPS handbook for what should or should not be described.
    \end{itemize}

\end{enumerate}

\end{document}